\documentclass[preprint,12pt,authoryear]{elsarticle}

\usepackage[a4paper,
            bindingoffset=0.2in,
            left=1in,
            right=1in,
            top=1in,
            bottom=1in,
            footskip=.25in]{geometry}

\usepackage{array}
\usepackage{booktabs}
\usepackage{multirow}
\usepackage{siunitx}
\usepackage{color,soul}
\usepackage{caption}
\usepackage{subcaption}
\usepackage{array}

\usepackage[utf8]{inputenc}
\usepackage{mathtools, nccmath}

\usepackage{amsmath}
\usepackage{amssymb}
\usepackage[mathlines]{lineno}
\usepackage{lineno}

\usepackage{algorithm} 
\usepackage{algpseudocode}

\usepackage[figurename=Fig.]{caption}

\DeclareCaptionLabelFormat{custom}{\textbf{#1 #2}}
\DeclareCaptionLabelSeparator{custom}{. }
\DeclareCaptionFormat{custom}{#1#2 #3}
\begin{document}

\begin{frontmatter}

\title{Deep-Learning Framework for Optimal Selection of Soil Sampling Sites}

\author[inst1]{Tan-Hanh Pham}
\author[inst2]{Praneel Acharya}
\author[inst3]{Sravanthi Bachina} 
\author[inst3]{Kristopher Osterloh}

\author{Kim-Doang Nguyen\corref{cor1}\fnref{inst1}}
\ead{knguyen@fit.edu}
\cortext[cor1]{Corresponding author}

\affiliation[inst1]{organization={Department of Mechanical and Civil Engineering},
            addressline={Florida Institute of Technology}, 
            country={USA}}

\affiliation[inst2]{organization={Department of Mechanical Engineering},
            addressline={Minnesota State University, Mankato}, 
            country={USA}}
            
\affiliation[inst3]{organization={Department of Agronomy, Horticulture \& Plant Science},
            addressline={South Dakota State University}, 
            country={USA}}

\begin{abstract}
Soil sampling is one of the most fundamental agricultural processes. The selection of soil sampling locations within a field plays a crucial role in soil-health analysis. The current soil-sampling techniques produce random samples across a field, which are not representative of the field conditions. This work leverages the recent advancements of deep learning in image processing for finding optimal locations that present the important characteristics of a field. The data for training are collected at different fields in local farms with five features: aspect, flow accumulation, slope, NDVI (normalized difference vegetation index), and yield. The soil sampling dataset is challenging because of the unbalance between the number of pixels that represent soil-sampling spots and the number of pixels that represent the background area. In this work, we approach the problem with two methods, the first approach involves utilizing a state-of-the-art model with the convolutional neural network (CNN) backbone, while the second is to innovate a deep-learning design grounded in the concepts of transformer and self-attention. Our framework is constructed with an encoder-decoder architecture with the self-attention mechanism as the backbone. In the encoder, the self-attention mechanism is the key feature extractor, which produces feature maps. In the decoder, we introduce atrous convolution networks to concatenate, fuse the extracted features, and then export the optimal locations for soil sampling. Currently, the model has achieved impressive results on the testing dataset, with a mean accuracy of 99.52\%, a mean Intersection over Union (IoU) of 57.35\%, and a mean Dice Coefficient of 71.47\%, while the performance metrics of the state-of-the-art CNN-based model are 66.08\%, 3.85\%, and 1.98\%, respectively. This indicates that our proposed model outperforms the CNN-based method on the soil-sampling dataset. To the best of our knowledge, our work is the first to provide a soil-sampling dataset with multiple attributes and leverage deep learning techniques to enable the automatic selection of soil-sampling sites. This work lays a foundation for novel applications of data science and machine-learning technologies to solve other emerging agricultural problems.
\end{abstract}

\begin{graphicalabstract}

\end{graphicalabstract}

\begin{highlights}
\item The first soil sampling site selection pipeline utilizing deep learning techniques has been developed so far.
\item The tool enables consistently collecting soil samples at the optimal locations under uncertain and variable conditions across different fields.
\item Developing a deep learning model to adapt multi-input images.
\item Mean accuracy, mean Intersection over Union, and mean Dice Coefficient on the testing dataset are 99.52${\%}$, 57.35${\%}$, and 71.47${\%}$, respectively.
\item We designed a deep-learning model with the self-attention mechanism as the backbone that outperforms the state-of-the-art CNN-based segmentation model.
\end{highlights}

\begin{keyword}
Deep learning\sep binary segmentation\sep self-attention\sep class imbalanced segmentation\sep soil sampling\sep precision agriculture
\end{keyword}

\end{frontmatter}


\section{Introduction}
\label{sec:intro}
\subsection{Background and motivation}
Soil is the foundation of agriculture, nurturing the seeds of civilizations since the beginning of humanity. Soil provides the essential nutrients and minerals that nourish the plants. Humans as well as most other species rely on this resource for food to sustain lives. Therefore, sustainable land management practices are crucial for maintaining a delicate balance in ecosystems, supporting the survival of most species, and promoting ecological stability. To achieve this, soil sampling is one of the most fundamental practices that enables producers to analyze and determine soil health, increase fertilizer use efficiency, and reduce agricultural runoff \citep{hodgson1978soil, tan1995soil, carter2007soil}.

The collection of soil samples from multiple locations within a field offers valuable insights into the physical, chemical, and biological properties of soil \citep{carter2007soil}. These properties are critical for various soil functions, including enhancing soil structure, providing physical support, facilitating nutrient cycling, and enabling water transport, all of which have direct impacts on crop yields \citep{brady2008nature}. Consequently, conducting a comprehensive soil analysis becomes essential in order to gather information about these properties. Soil health serves as valuable input for optimizing nutrient-utilization efficiency and agricultural productivity \citep{wolf1999fertile}. To ensure accurate soil analysis results, it is important to follow proper soil sampling protocols and collect representative samples from multiple locations within the field or area being tested \citep{tan1995soil}.

Understanding the spatial variability of soil properties typically requires an extensive degree of scientific knowledge and training, and a thorough amount of legacy soil data for the field of interest. Topography is one such key soil-forming factor and has a significant influence on a wide array of soil physical and chemical properties \citep{ceddia2009topography}. Terrain attributes, such as slope, aspect, and curvature, have strong relationships with many soil properties \citep{miller2015digital} including water hydraulic conductivity \citep{herbst2006geostatistical, sarapatka2018varying}, aggregate stability \citep{zadorova2011influence, jakvsik2015soil}, and soil organic carbon \citep{stolt2010insights}, which are strongly associated with productivity. This information on spatial heterogeneity of soil properties is a prerequisite for sustainable resource use, site-specific management of plant nutrients \citep{plant2001site}, and designing appropriate soil sampling schemes \citep{kerry2010sampling}.

\begin{figure}[h]
\begin{center} 
\includegraphics[width= 0.8\textwidth]{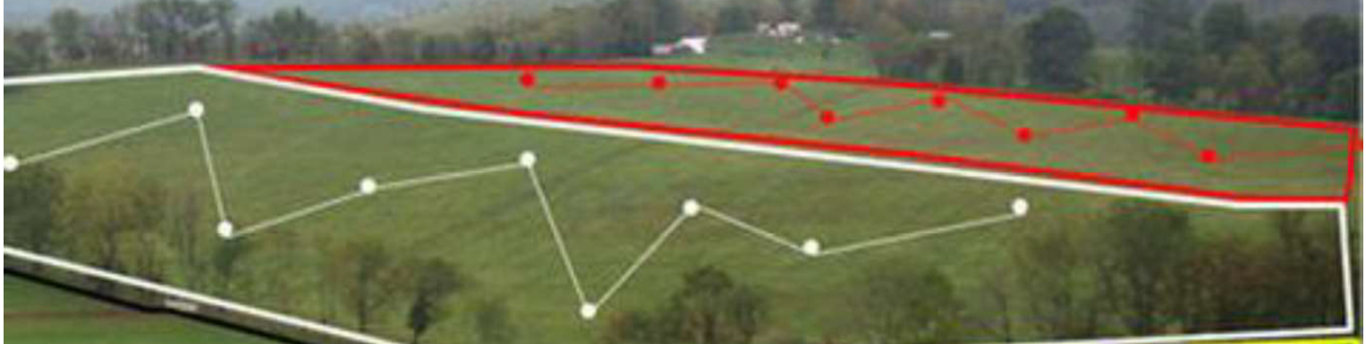}
\end{center}
\caption{Soil sampling zigzag pattern: The red pattern is more regular since the terrain is flat, while the white pattern is selected based on uneven terrain properties.}
\label{fig.field} 
\end{figure}

In current agricultural practices, soil samples are collected and sent to laboratories for analysis of soil properties. There are scientific methods that involve pooling samples based on landscape positions, historical data, and knowledge of how nutrients move in soil due to variations in soil properties \citep{rowell2014soil}, for example, composite sampling, cluster sampling, and stratified sampling. However, these methods can be time-consuming, expensive, complex, and may have limited precision and spatial coverage \citep{dane2020methods}. While aggregate soil sampling can provide an overview of overall field conditions, it is not suitable for precision agriculture techniques as it does not account for the dynamic complexity of soil properties \citep{liu2009spatio}. Figure \ref{fig.field} illustrates an example of standard zigzag sampling patterns for accurate soil analysis \citep{dane2020methods}. 

Most producers nowadays do not follow those scientific procedures as they can be complex and vary from field to field. Instead, they usually collect soil samples from multiple random locations across a field. There is no assurance of the extent to which the samples actually represent the field's soil health \citep{oliver2010integrating}. As a result, pooled soil samples may not be representative of the actual variation in soil properties across the field. This leads to the problem that there are no reliable methods for farmers to accurately utilize knowledge of soil variation with their precision agriculture technologies \citep{keith1991environmental}. Therefore, there is an urgent need for an automated tool that helps producers identify which spots are optimal for sampling and can be pooled to get accurate soil analysis results \citep{lawrence2020guiding}. To fill this critical need, this paper develops a deep-learning tool that outputs optimal locations in a given field to collect soil samples for analysis. Our central hypothesis is that advanced deep-learning techniques to analyze and refine landscape data will enable precise and reliable recommendations of optimal soil sampling spots. The rationale is that the deep learning tool extracts meanings from landscape data and human-labeled data to train multi-layer neural networks that increase the reliability and accuracy of sampling-location selection \citep{najafabadi2015deep}. Actual field tests including the geographical features depicted in Fig. \ref{fig.field} are considered with experimental results collected from local farms. The paper demonstrates that the developed framework produces consistent results under uncertain variable conditions from field to field. 

\subsection{Literature review}
Machine learning techniques, including K-means, support vector machines (SVM), and artificial neural networks (ANN), have proved their efficacy in extracting patterns from spatial data \citep{alzubi2018machine}. Recently, a subset of machine learning, deep learning, has been rapidly growing with many applications in autonomous vehicle \citep{kuutti2020survey}, healthcare \citep{miotto2018deep}, and agriculture \citep{kamilaris2018deep}. Deep learning algorithms are designed to automatically learn and extract meaningful patterns and representations from large amounts of data. These algorithms are organized into multiple layers of interconnected artificial neurons, known as artificial neural networks. Each layer receives input from the previous layer and performs computations to transform the data. The final layer produces the output or prediction depending on the specific tasks. Deep learning uses the backpropagation algorithm to discover the complex structure of big data by adjusting its internal parameters used to model the relationship of hierarchical layers \citep{lecun2015deep}.

The use of convolution neural networks (CNNs), a type of deep neural networks, to extract hierarchical features has shown success in various computer vision tasks, including image classification, object detection, and image segmentation \citep{lecun1995convolutional, lecun1998gradient}. At the core of a CNN are convolutional layers, which extract local features from an input image by applying convolution operations. Convolution involves sliding a small filter (also known as a kernel) over the input image, performing element-wise multiplication, and summing the results to produce local receptive features to various objects in the image. Each unit of a layer inherits features from a set of neighborhood units in the previous layer, and the outputs of such units constitute a feature map. The filter's weights are learned during the training process, allowing the network to automatically discover relevant patterns and features. Some of the most popular CNN architectures include AlexNet \citep{alexnet}, VGGNet \citep{vgg}, ResNet \citep{resnet}, Unet \citep{unet}, and SegNet \citep{segnet}. Popular algorithms used for object detection include R-CNN \citep{RCNN}, Faster R-CNN \citep{ren2015faster}, YOLO family \citep{yolo}, and SSD \citep{ssd}, while Fully Convolutional Networks (FCN) \citep{long2015fully}, Mask RCNN \citep{he2017mask}, DeepLab family \citep{chen2017rethinking} algorithms are well known for image segmentation. 

The self-attention method has achieved great success in the field of natural language processing \citep{vaswani2017attention}. Recently, it has also shown potential in computer vision tasks including object detection, classification, and segmentation \citep{carion2020end, dosovitskiy2020image, xie2021segformer}. The self-attention mechanism is utilized as spatial attention to extract features of an image as convolution layers. Unlike the localization of the CNNs, the advantage of self-attention is the ability to understand receptive fields globally \citep{wang2018non, vaswani2017attention}. The response of each unit of a layer is a weighted sum of the features in the previous layers. 

Machine learning and deep learning also found compelling applications in agriculture to improve food production, including plant disease detection \citep{li2022multi}, pest recognition \citep{wang2020common}, and crop identification \citep{pandey2022intelligent}. For example, work in \cite{naimi2021spatial} used machine learning methods to predict the soil surface properties. Deep-learning techniques were also leveraged in \cite{acharya2022ai} and \cite{acharya2023deep} to detect and track droplets and estimate droplet size, velocity, and spray angles. These parameters are crucial in determining the characteristics of crop spray systems and their optimal designs.

Precision agriculture has improved in spatial resolution, accuracy, and reliability in many applications by leveraging recent technological developments. However, soil sampling is one of the very few areas in precision agriculture that have not significantly advanced for decades \citep{kno18}. Soil survey data is available for most of the United States but is mapped at a scale that is too coarse to be useful in precision agriculture. As a result, current soil analysis results do not reflect the true distribution of soil properties over a field and undermine the efficacy of the subsequent agricultural processes. According to \cite{hengl2017soil}, soil sampling sites affect the reliability of the nutrient prediction, and it is critical to optimize the sampling locations taking into account geological and pedological features of the field. There has not been any substantial research on leveraging machine learning to improve soil sampling practices.

\subsection{Contribution}
In this paper, we leverage recent advancements in machine learning and computer vision to create an efficient tool for soil sampling site selection. The goal can be achieved by formulating a deep learning model capable of predicting soil sampling sites with accuracy at around 90${\%}$. The tool enables a significant improvement in soil sampling and analysis, which lead to a better understanding of soil health. 

The key contributions of this paper are articulated as follows:
\begin{itemize}
\item We have created a deep-learning framework that uses landscape data to autonomously learn, analyze, and improve the selection of soil sampling locations. To the best of our knowledge, this is the first work that applies deep-learning techniques to soil sampling site selection.
\item The tool facilitates the consistent collection of soil samples at optimal locations despite the variable and uncertain conditions encountered across different fields. It is thoroughly trained with experimental data collected from local farms and annotated by pedologists.
\item We have designed and implemented a set of performance metrics to assess the success rate of the newly developed tool for optimal sampling site prediction.
\item The developed deep-learning model is capable of accommodating multiple layers of spatial data. Specifically, the input data are composed of five layers: aspect, flow accumulation, slope,
normalized difference vegetation index (NDVI), and yield.
\end{itemize}

The rest of the paper is organized as follows. Section \ref{sec.2} describes the soil-sampling data collection and data pre-processing. Section \ref{sec.3} discusses the methodology of the soil sampling tool. This section explains how the tool takes images in as inputs, extracts patterns from the inputs, and then outputs soil sampling locations. In this section, we provide a detailed algorithm that describes how the deep learning model behind the tool works. Section \ref{sec.metric} elaborates on a set of metrics used to quantify the model performance. The result and discussion of the soil-sampling site selection tool are shown in Section \ref{sec.resultanddiscussion}. Section \ref{sec.6} concludes the report, proposes potential strategies to address the existing challenges, and outlines future directions for further research.

\section{Soil sampling data collection and annotation}
\label{sec.2}
\subsection{Data Collection}
In order to build a deep learning tool, soil sampling data were collected in Aurora and Davison counties, South Dakota. A total of fourteen fields, ranging from 150 to 200 hectares each, were selected. These fields encompassed diverse landscapes and soil properties and were under the management of local farmers. For training and testing the model, each field was characterized by five attributes: aspect, flow accumulation, slope, yield, and normalized difference vegetation index (NDVI). The corresponding ground truths were also annotated by the pedologists in our team. Digital Elevation Models (DEM) for each field were downloaded from the LiDAR (Light Detection and Ranging) dataset in South Dakota Geological Survey with a spatial resolution of one meter (https://www.sdgs.usd.edu/).

\begin{figure}[h]
\begin{center} 
\includegraphics[width= 0.8\textwidth]{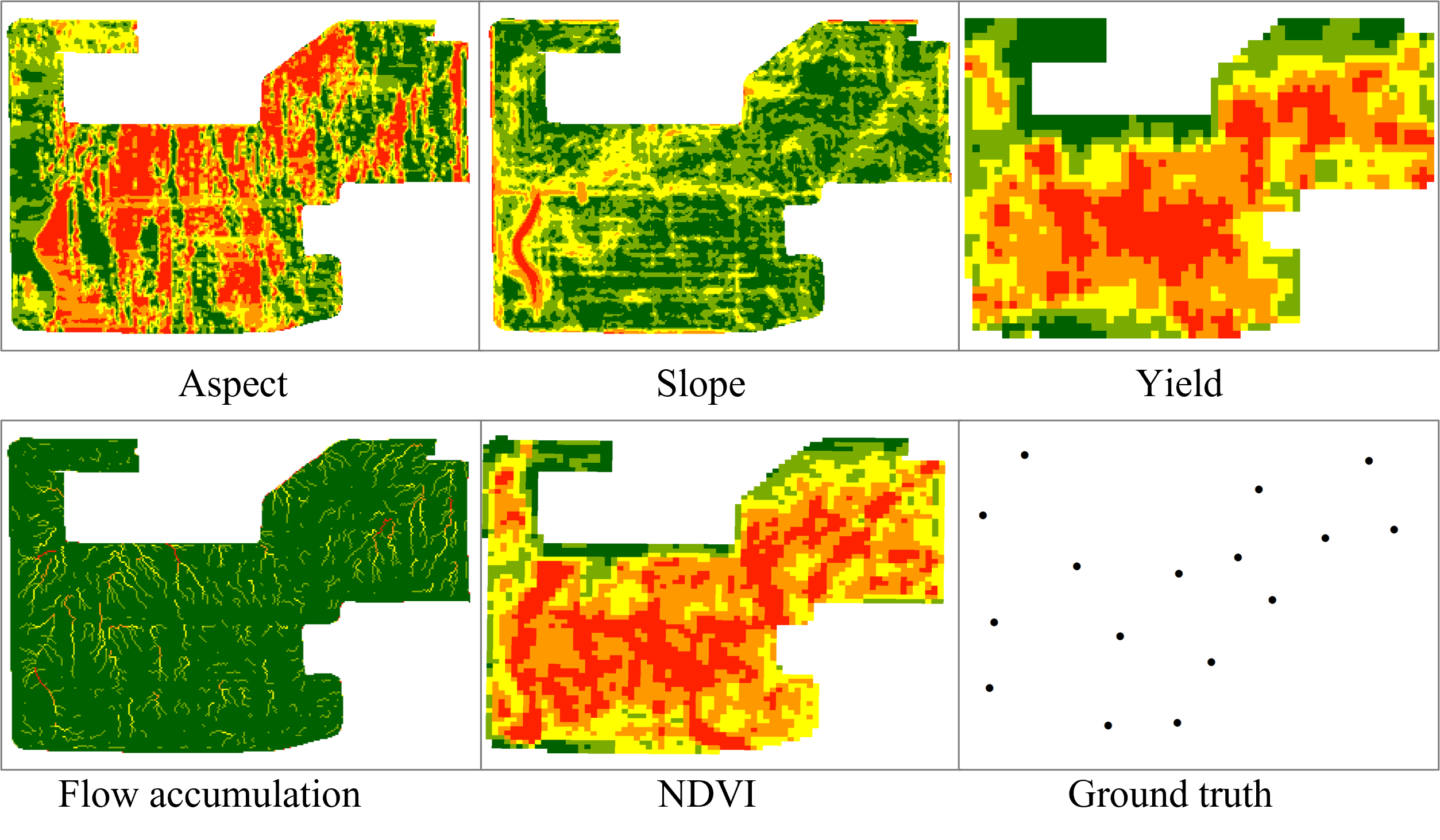}
\end{center}
\caption{A set of raw datasets representing terrain and yield attributes of a field.}
\label{fig.raw_dataset} 
\end{figure}

Terrain attributes (slope and aspect) and hydrological attributes (flow accumulation) were processed by using ArcMap 10.8 (Esri, ArcGIS). 
Soil properties like water holding capacity, soil texture, soil organic matter, and Cation Exchange Capacity (CEC) were obtained from the SSURGO dataset \citep{SSURGO}. 
For this research, Sentinel 2A imagery was downloaded from the Copernicus Open Hub website with a spatial resolution of 10 meters which is further used to obtain NDVI values for each field. Yield data were obtained by Field View Plus (Climate LLC) which was further cleaned and processed by using SMS Ag software (Ag Leader, SMS Advanced). For model training, terrain and yield attributes including slope, aspect, flow accumulation, yield, and NDVI data are used as features or independent variables. The corresponding ground truths soil sampling sites of these set data were labeled, which represent the characteristics of the field as shown in Fig. \ref{fig.raw_dataset}.

\subsection{Data pre-processing and augmentation}
\begin{figure}[h]
\begin{center} 
\includegraphics[width= 0.8\textwidth]{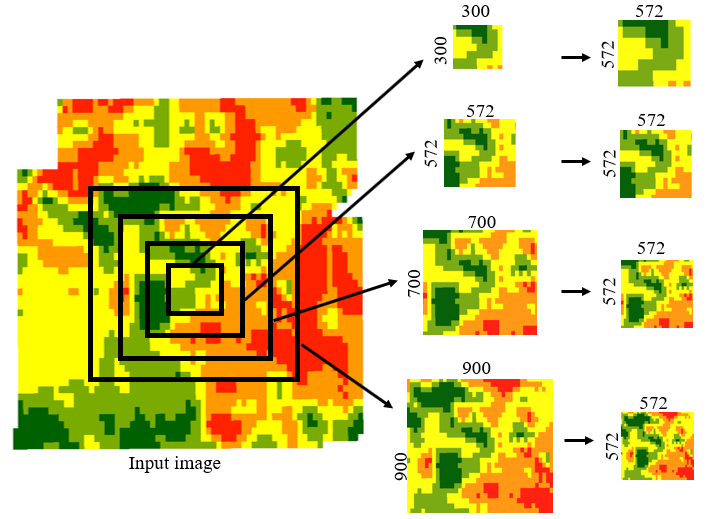}
\end{center}
\caption{Graphical representation of cropping a square image from the center of input image with sizes of 900, 700, 572, and 300 pixels, and resizing them to an image size of 572 by 572 pixels. }
\label{fig.image_cop} 
\end{figure}

\begin{figure}[h]
  \centering
  \begin{subfigure}{0.45\textwidth}
    \centering
    \includegraphics[width=\linewidth]{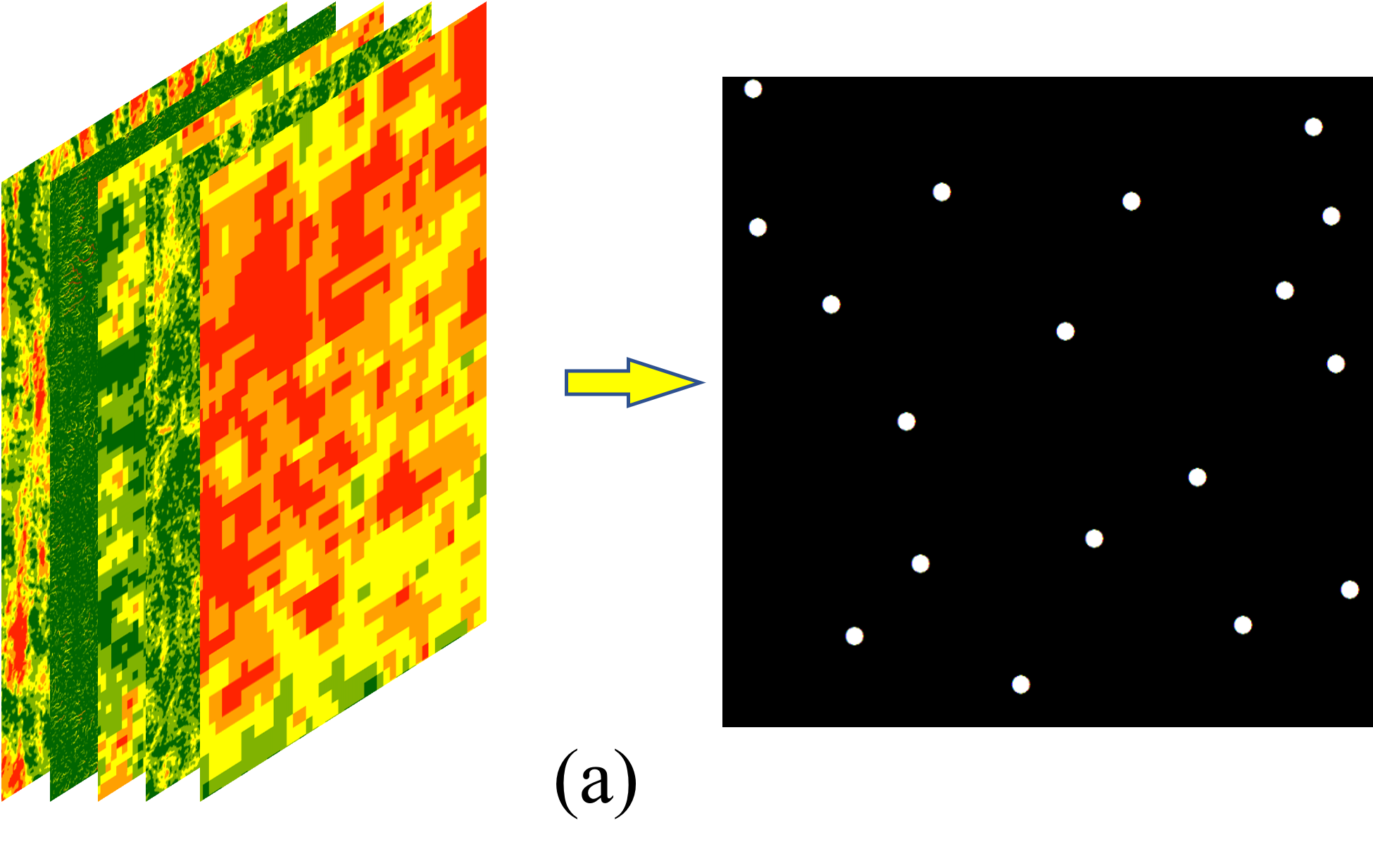}
    \label{fig.data1}
  \end{subfigure}
\hfill
  \begin{subfigure}{0.45\textwidth}
    \centering
    \includegraphics[width=\linewidth]{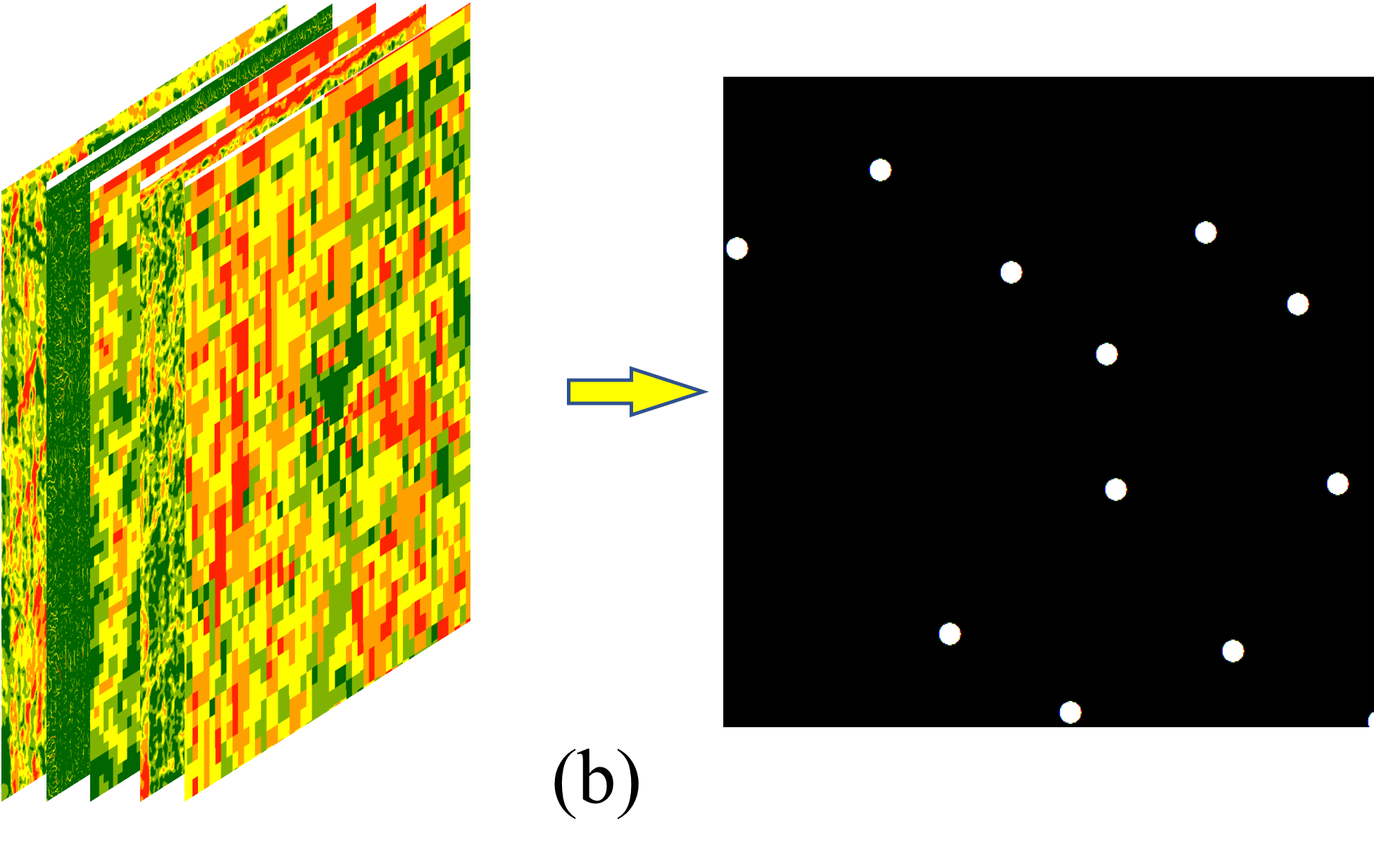}
    \label{fig.data2}
  \end{subfigure}
  \caption{Representation of soil sampling dataset with field attributes stacked as an input and respective ground truth. Each input represents the portion of a collected field as shown in Fig. \ref{fig.raw_dataset}. In each set, the input images are on the left, and the corresponding ground truth is on the right.}
  \label{fig:dataset}
\end{figure}

The data for each field were organized into a stack of five images corresponding to their key attributes: aspect, flow accumulation, slope, yield, and NDVI. These input images have various sizes and shapes depending on the field, with an average height of 800 pixels and width of 1100 pixels. In order to make the input data uniform and increase the size of the training dataset, we augmented the data by implementing the following steps: 
\begin{itemize}
\item Cropping multiple square images from the center of the input image with sizes of $900\times 900$, $700\times 700$, $572\times 572$, and $300\times 300$ pixels.
\item Resizing them to an image with the size of $572 \times 572$ pixels. This step is visually illustrated in Fig. \ref{fig.image_cop}.
\item Ten random crops are taken from the input image, each with a size of $572 \times 572$ pixels.
\item Each input image undergoes twenty random rotations to angles between five and sixty degrees, and each output from these rotations is resized to the size of $572 \times 572$ pixels.
\item Images collected after image augmentation are grouped into three sets: training set (1455 samples), validation set (180 samples), and testing set (185 samples). The representatives of the data are shown in Fig. \ref{fig:dataset}.
 \end{itemize}

The ground truth binary image used in training consists of two classes: foreground (white pixels in Fig. \ref{fig:dataset}), representing the desired soil sampling locations, and background (black pixels in Fig. \ref{fig:dataset}), containing everything else. The ratio of background to foreground in all ground truth images is illustrated in Table \ref{tab:ratio_GT}. Since the ratio of the two classes (total number of black pixels in a given image over the total number of white pixels) on average is 146.368, the data is highly imbalanced. Training the model becomes tremendously challenging due to the uneven distribution of these classes of pixels. This is because, it is possible for the model to predict everything as background and achieve a near-zero loss value, despite failing to accurately capture the true foreground pixels. To address this challenge, the next section will discuss our method grounded in the concepts of transformers and self-attention.

\begin{table}[h]
\centering
\caption{Ratio of foreground pixels to background pixels in all ground truth images.}
\label{tab:ratio_GT}
\begin{tabular}{@{}cccccc@{}}
\toprule
Minimum & Maximum & Median & Average & Standard deviation \\ \midrule
33.429  & 783.613   & 124.816   & 146.368   & 100.150                       \\ \bottomrule
\end{tabular}
\end{table}

\section{Methodology}
\label{sec.3}
\subsection{Overview of the method}

The research goal of this work is to develop a tool that analyzes the characteristics of a given field and assists farmers in selecting optimal soil sampling locations in the field. This goal is achieved via a deep-learning segmentation methodology that leverages recent advances including vision transformers and self-attentions. The soil sampling input dataset is unconventional compared to other computer vision tasks as it comprises multiple layers of input data representing distinct attributes of a field: aspect, flow accumulation, slope, NDVI, and yield. The model's output is a binary image that indicates the optimal locations for collecting soil samples. Therefore, we formulate the soil-sampling site selection as a binary segmentation problem with white pixels representing the proposed sampling locations and black pixels as the background.

\begin{figure}[h]
\begin{center}
\includegraphics[width= 1\textwidth]{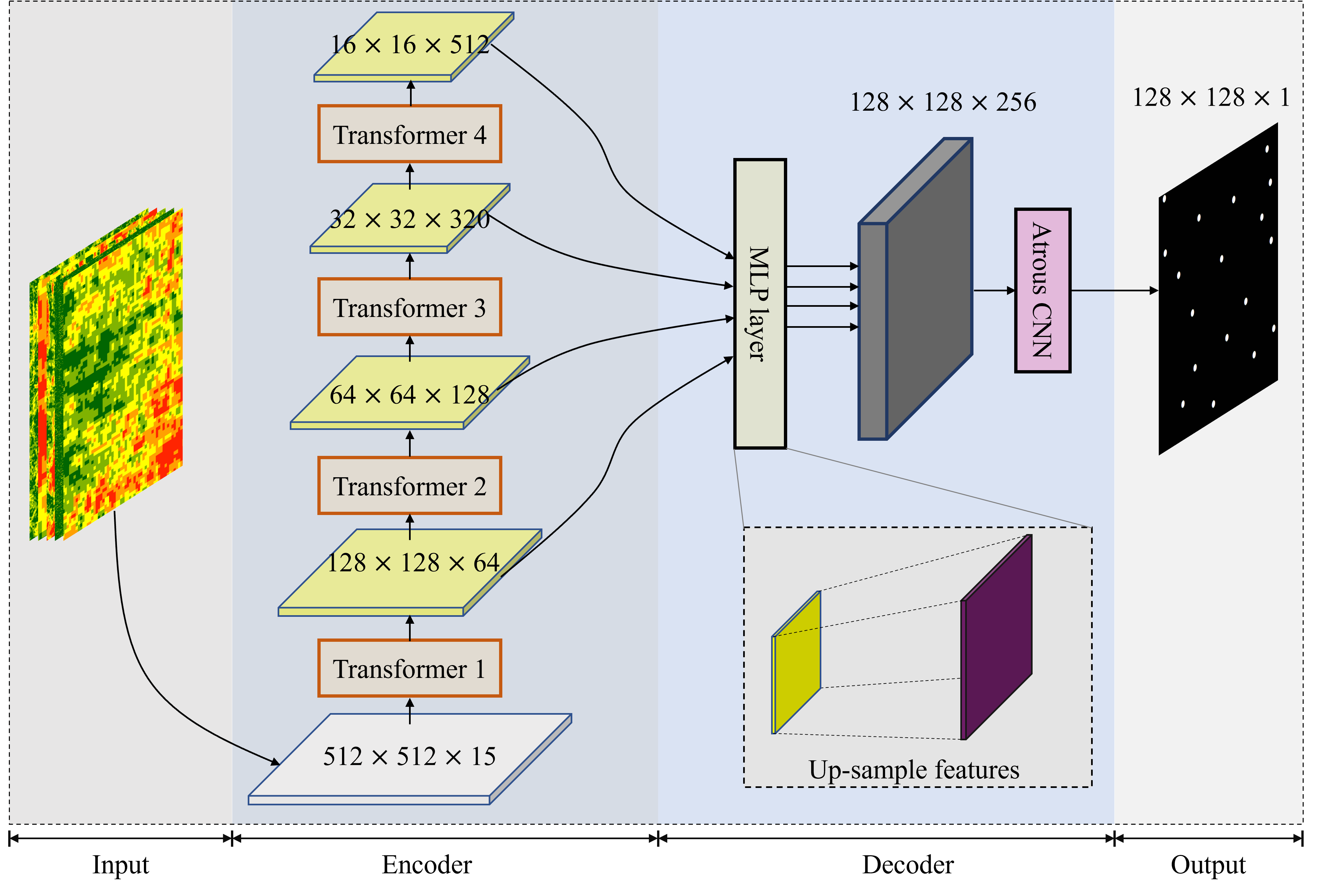}
\end{center}
\caption{Soil sampling tool with the combination of SegFormer model and image processing algorithms.}
\label{fig.workflow} 
\end{figure} 

To solve the problem, we leverage the transformer-based semantic segmentation architecture, called SegFormer, proposed in \cite{xie2021segformer}, integrated with our atrous convolution networks, to design a tool, a semantic segmentation model, that takes a stack of fifteen layers of soil input data as input and outputs a binary mask with recommended soil sampling locations. The semantic segmentation model primarily employs self-attention as extractors to model the relationships between different patches and extract features of input soil data. The model is constructed with an encoder and a decoder. The encoder generates high-resolution fine features to low-resolution coarse features. The input data are downsampled multiple times to extract hierarchical features in the transformer backbone. The decoder is responsible for upsampling these features back to the original image resolution, concatenating them, and then outputting predictions. This process is essential to ensure that the segmentation predictions have the same spatial dimensions as the input data. Specifically, the workflow of the machine-learning tool for soil sampling site selection is depicted in Fig.~\ref{fig.workflow}.

\begin{figure}[b]
\begin{center}
\includegraphics[width= 0.8\textwidth]{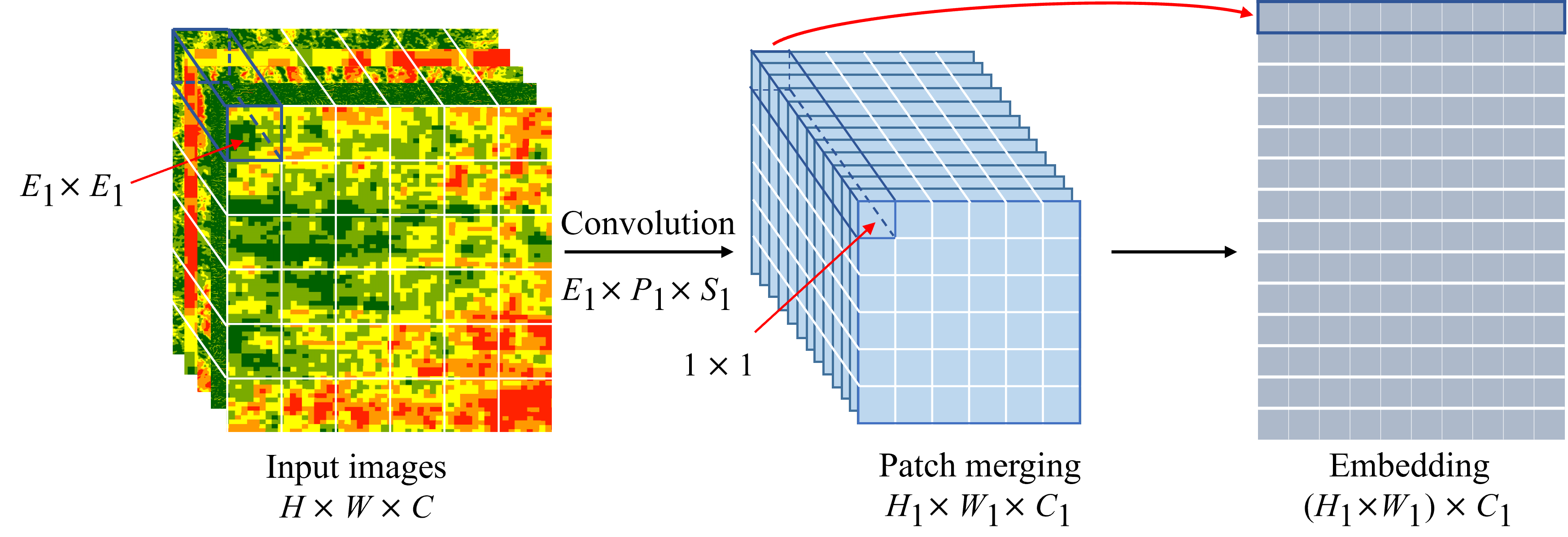}
\end{center}
\caption{The overlapped patch merging process used for shrinking input data and embedding that to sequences.}
\label{fig.overlap patch merging} 
\end{figure} 

\subsection{The encoder}
\subsubsection{Patch merging}
The encoder is designed with self-attention mechanisms at its core that enable elements in the stack of input soil data to focus on relevant information by computing attention scores with respect to other elements. The encoder efficiently captures long-range dependencies and contextual relationships, making it effective for sampling site selection since understanding global interactions is important. Although self-attention architecture has found many successes in natural language processing (NLP) \cite{vaswani2017attention}, our work is the first paper that integrates and implements self-attention operations for soil sampling data. 

Inspired by the effort to apply self-attention mechanism in computer vision tasks in \cite{dosovitskiy2020image}, this work first transforms the input dataset of fifteen layers into sequences (like words in NLP) via a patch embedding operation, and then using the self-attention mechanism to extract the relationship between the sequences. Figure \ref{fig.overlap patch merging} depicts the patch embedding process for the first Transformer block. In particular, an input dataset of size ($H$, $W$, $C$), in which $H$ is the height, $W$ is the width, and $C$ represents the number of channels, is partitioned into a number of patches with the patch size of $E_1\times E_1$, and then all elements in a patch are merged together. To do that, the input is fed through a convolutional layer with a kernel size equal to patching size ($E_1$), a stride (\emph{$S_1$}), a padding (\emph{$P_1$}) and a spatial dimension (\emph{$C_1$}). After this overlapped patch merging operation, the size of the input is shrunk to 

\begin{flalign}\label{eq.shrinking}
    \begin{aligned}
        H_1 & = \frac{(H - E_1 + 2P_1)}{S_1} + 1, ~~~\mbox{and} \\
        W_1 & = \frac{(W - E_1 + 2P_1)}{S_1} + 1. 
    \end{aligned}
\end{flalign}

This results in a stack of features of the size $H_1\times W_1\times C_1$, in which the dimensions of a patch are shrunk from ($E_1\times E_1$) to ($1 \times 1$). After that, the patches are embedded into linear sequences as the inputs of the self-attention mechanism. Particularly, Figure \ref{fig.overlap patch merging} shows the overlap patch merging of Transformer 1 while Figure \ref{fig.self-attention} illustrates the overall operation of a Transformer block and how patch merging fits into the self-attention.

\subsubsection{Self-attention}
The self-attention in Transformer 1 then takes the embedded sequences as input, and then generates three row-vectors for each sequence $s$: query (\emph{q}) of dimension $d_q$, key (\emph{k}) of dimension $d_k$, and value ($v$) of dimension $d_v$. These vectors are created by pre-multiplying the embedded sequences with learned-weight matrices $W^Q \in \mathbb{R}^{C_1\times d_q}$, $W^K \in \mathbb{R}^{C_1\times d_k}$, and $W^V \in \mathbb{R}^{C_1\times d_v}$, i.e. 
\begin{equation}
\label{eq.qkv}
q=sW^Q,~~~k=sW^K, ~~~v=sW^V
\end{equation}
In our algorithm, we set $C_1$ = $d_q$ = $d_k$ = $d_v$.

These learned-weight matrices contain parameters to be trained during the training process discussed at the end of this section. For a group of input sequences, we define a query matrix $Q \in \mathbb{R}^{N \times d_q}$, a key matrix $K \in \mathbb{R}^{N \times d_k}$, and a value matrix $V \in \mathbb{R}^{N \times d_v}$, where $N = H_1 \times W_1$. Let the query vector $q_i$ be the $i^{th}$ row of $Q$ and the corresponding key vector $k_j$ be the $j^{th}$ row of $K$, the score between them, $\hat{\alpha}_{i,j}$, can be calculated as the dot product:
\begin{equation}
\label{eq.attention ij}
\hat{\alpha}_{i,j} = q_ik_j^T, \forall{i, j} \in N.
\end{equation}

The attention weight $\alpha_i$, which presents how much a sequence $i^{th}$ is related to other sequences, is then calculated using the softmax function:
\begin{equation}
\label{eq.attention i}
\mbox{$\alpha$}_i = \mbox{softmax}\left(\frac{\hat{\alpha}_{i}}{\sqrt{d_k}}\right), \forall{i} \in N.
\end{equation}
Particularly, in the vector $\alpha_i$, $\alpha_{i,j}$ represents the relevance of the sequence $i^{th}$ to the sequence $j^{th}$. The softmax function takes a vector of real numbers as input and transforms them into a probability distribution over multiple classes, where the sum of the probabilities is equal to 1. The reason for dividing by $\sqrt{d_k}$ is to achieve more stable gradients.

Once the attention weights are calculated, the attention of the sequence $i^{th}$ is computed as: 
\begin{equation}
\label{eq. single attention}
\mbox{Attention}_i(q_i, K, V) = \alpha_iV, \forall{i} \in N.
\end{equation}
During this process, the computational complexity of the attention is $O(N^2)$, however, we can reduce it \emph{R} times by applying the sequence reduction process for the key matrix $K \in \mathbb{R}^{N/R \times d_k}$ and $V \in \mathbb{R}^{N/R \times d_v}$. So, the complexity becomes $O(\frac{N^2}{R})$. 

\begin{figure}[H]
\begin{center}
\includegraphics[width= 1\textwidth]{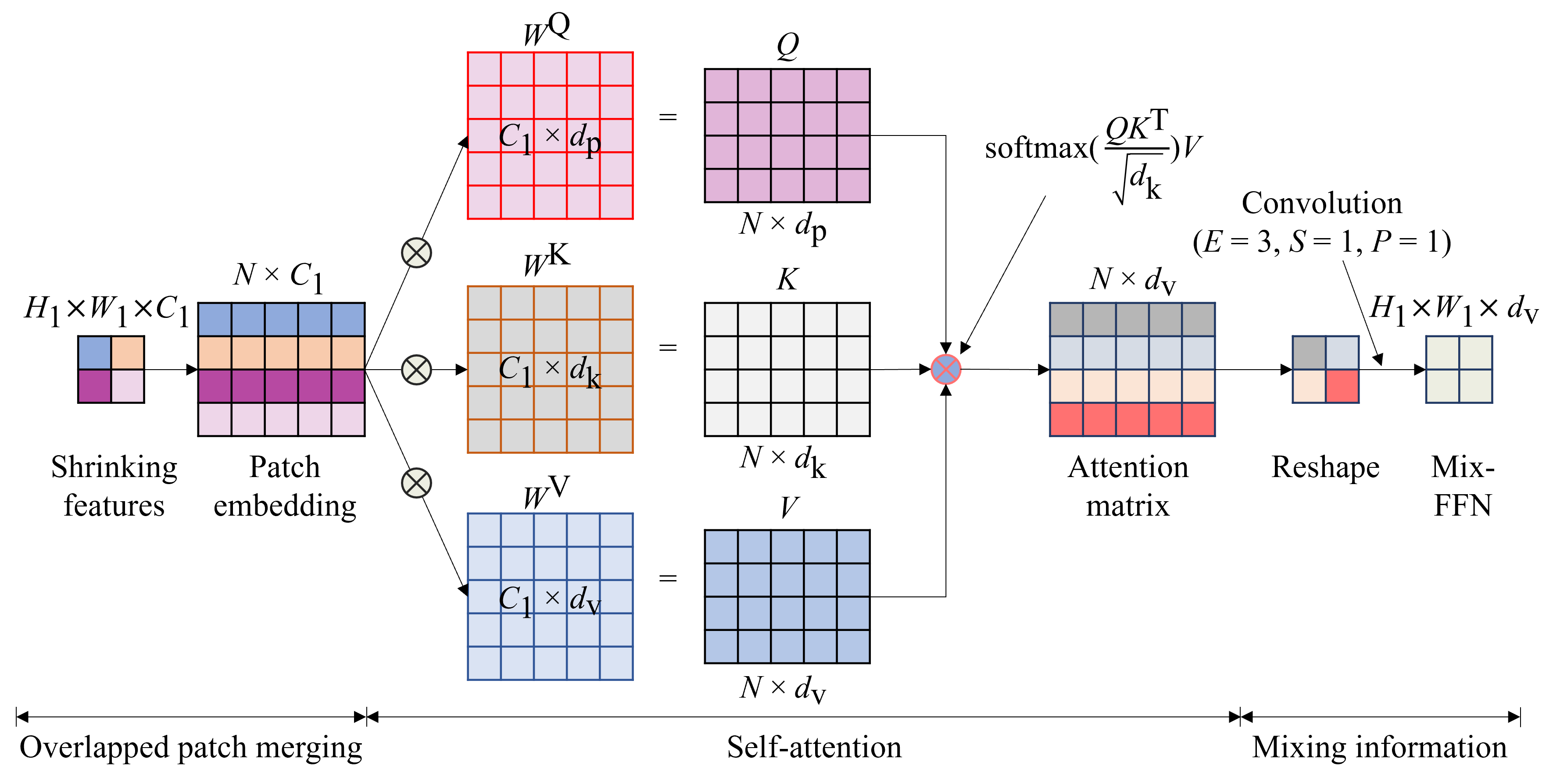}
\end{center}
\caption{Operation within Transformer 1 with one head attention.}
\label{fig.self-attention} 
\end{figure}

Instead of using a single attention function, multi-head attention is used to process the soil data. The use of multi-head self-attention can efficiently bring the model to parallel computing and empowers different attention heads to focus on different patterns in the input sequences. Similar to the derivation above for an individual sequence, the attention matrix for $N$ sequences over a head is calculated as follows:
\begin{equation}
\label{eq. self attention}
\mbox{Attention}(Q, K, V) = \mbox{softmax}(\frac{QK^T}{\sqrt{d_h}})V.
\end{equation}
Equation \eqref{eq. self attention} presents the attention matrix in head attention, therefore, the final attention matrix is computed by concatenating all multi-head attentions in the embedding dimension, i.e.
\begin{equation}
\label{eq. multi attention}
\mbox{MultiHead}(Q, K, V) = \mbox{Concat}(\mbox{head}_1, ..., \mbox{head}_h)W^O.
\end{equation}
where $\mbox{head}_e = \mbox{Attention}(QW_e^Q, KW_e^K, VW_e^V)$; $W_e^Q \in \mathbb{R}^{N \times d_h}$, $W_e^K \in \mathbb{R}^{N/R \times d_h}$, and $W_e^V \in \mathbb{R}^{N/R \times d_h}$ are parameter matrices over a head, and $W^O \in \mathbb{R}^{N \times d_k}$ is the total parameter matrix. Figure~\ref{fig.self-attention} depicts the role of self-attention in Transformer 1.

\subsubsection{Feed Forward}
The output of the self-attention mechanism serves as the input for the feed-forward network (Mix-FFN) within a Transformer block, as illustrated by Fig. \ref{fig.self-attention}. In the Mix-FFN, the output from self-attention is reshaped to match the input dimensions and then mixed using CNNs. The process is repeated $L$ times before passing to the next Transformer. Similar to Transformer 1, the feature extraction in the other Transformers can be constructed in the same way but with different parameter values.

For example, in Transformer 1, the values of \emph{$E_1$}, \emph{$S_1$}, \emph{$P_1$}, and \emph{$C_1$} are 7, 4, 3, and 64, respectively. The input soil dataset, with dimensions of 512 pixels by 512 pixels by 15 channels, is divided into 16,384 patches through overlapped patch merging. These patches are then transformed into sequences with a length of 64. Subsequently, the sequences undergo the self-attention mechanism and are transformed into different sequence outputs. The outputs are reshaped and mixed using a convolutional network within the Mix-FFN, and this process is repeated three times. Following this, the output sequences' dimensions are ($128 \times 128 \times 64$). 

Similar to Transformer 1, we used the values of \emph{$E_i$} = 3, \emph{$S_i$} = 2, \emph{$P_i$} = 1, \emph{$C_i$} = 128, 320, and 512, and $L_i$ = 3, 18, and 3, for $i=2,$ $3,$ and $4$, respectively, for the other Transformer blocks. Each Transformer block further reduces the resolution of the input data as ($H_i$, $W_i$) while hierarchically increasing the spatial dimension ($C_i$) in the encoder. There are four Transformer blocks in the encoder, so the sizes of input data will become $(\frac{H}{4}\times \frac{W}{4}\times C_1)$, $(\frac{H}{8}\times \frac{W}{8}\times C_2)$, $(\frac{H}{16}\times \frac{W}{16}\times C_3)$, $(\frac{H}{32}\times \frac{W}{32}\times C_4)$. 
In particular, the input has dimensions of ($512 \times 512 \times 15$), after passing through the sequence of Transformer blocks, the sizes of the data are ($128 \times 128 \times 64$), ($64 \times 64 \times 128$), ($32 \times 32 \times 320$), and then ($16 \times 16 \times 512$). 

\subsection{Decoder}
In the decoder, all outputs from the hierarchical Transformer blocks are up-sampled from their respective size to $(\frac{H}{4}\times \frac{W}{4}\times 768)$ by using a multi-layer perceptron (MLP) and a linear interpolation function. The MLP is a linear transformation function that maps the input features $C_i$ to incoming outputs $C_h$ while the interpolation function upscales the feature's size linearly as follows:
\begin{equation}
\label{eq.upsample}
\hat{F_i} =\mbox{Upsample}_{\left(\frac{H}{4}\times \frac{W}{4}\right)}\left(\mbox{Linear}_{(C_i, C_h)}(F_i)\right), \forall{i}.
\end{equation}
Then, these up-sampled features $\hat{F_i}$, from Transformer $i^{th}$, are concatenated and fused together by using an atrous convolution layer as:
\begin{equation}
\label{eq.linear2}
F = \mbox{Conv}_{(4C_h, C_m)}\left(\mbox{Concat}(\hat{F_i})\right), \forall{i}.
\end{equation}
Here, $C_m$ is the number of the output features of this process.

\begin{figure}[H]
\begin{center}
\includegraphics[width= 0.8\textwidth]{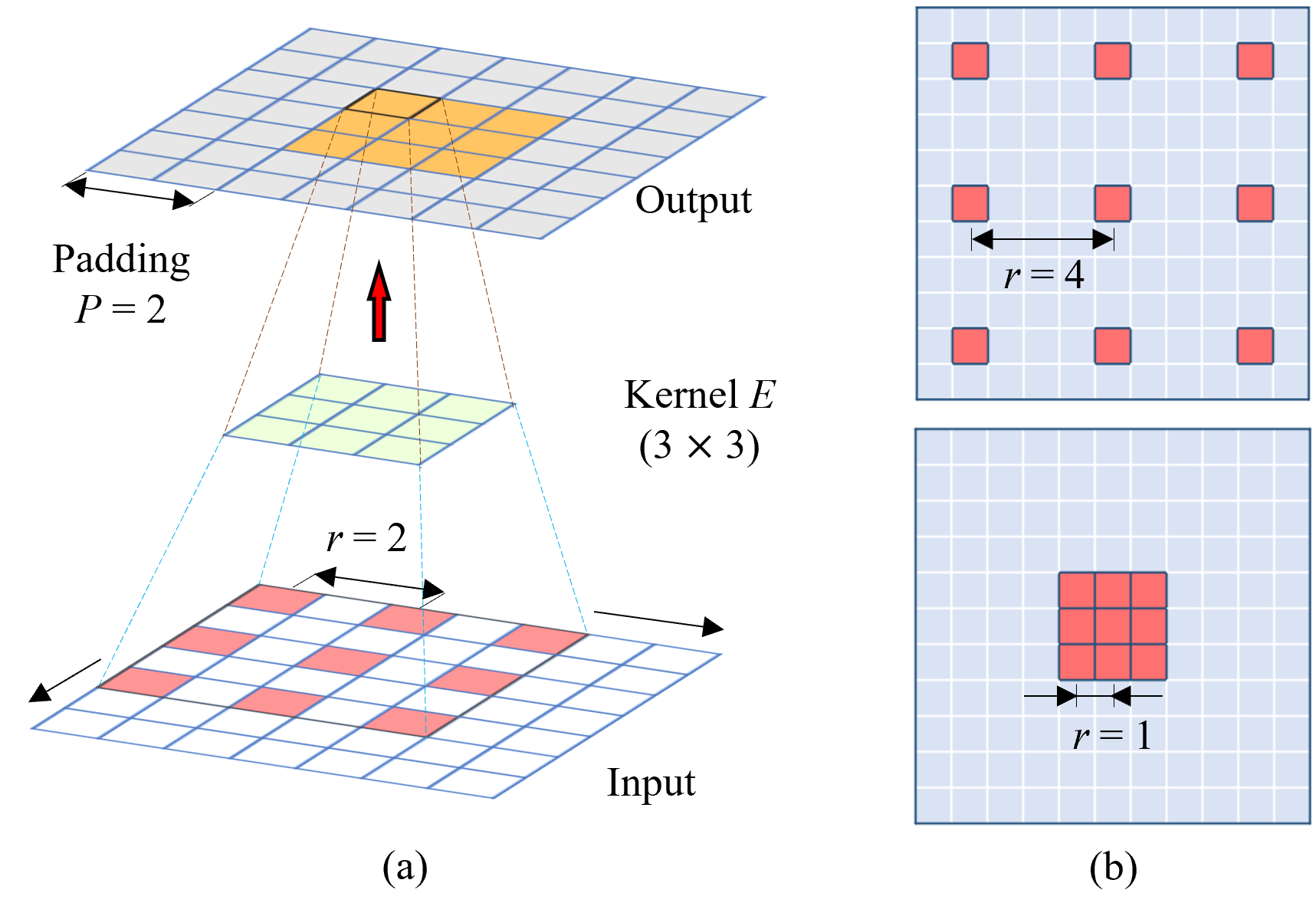}
\end{center}
\caption{(a) Atrous convolution with the dilated rate $r$ = 2, kernel size $E$ = 3, stride $S$ = 1 and padding size $P$ = 2. (b) Visualization of receptive fields when using atrous convolution with $r$ = 1 and $r$ = 4.}
\label{fig.atrous} 
\end{figure}

In this work, we intentionally use atrous convolution networks (dilated convolution) because this helps capture the global context by considering a wider region around each pixel. The atrous convolution operation, illustrated in Fig. \ref{fig.atrous}, allows the convolutional layer to have a larger effective receptive field without increasing the number of parameters or the computational cost. This is achieved by adding "holes" with the dilated rate $r$ in the filter, which leads to an expansion of the receptive field. A larger dilated rate means more gaps and a larger receptive field, which allows the network to capture broader context information. Typically, when the dilated rate $r$ is one, which is the same as the traditional CNN.

In particular, to keep the same dimensions as the input layers, we used atrous convolution with $r$ = 2, $E$ = 3, $S$ = 1, $P$ = 2, and output layers $C_m$ = 256. The output dimensions of atrous convolution, ($H_{out}$, $W_{out}$), are calculated as 

\begin{flalign}\label{eq.atrous}
    \begin{aligned}
        H_{out} & = \frac{H_{in} +2P - r(E - 1) - 1}{S} +1, ~~~\mbox{and} \\
        W_{out} & = \frac{W_{in} +2P - r(E - 1) - 1}{S} +1. 
    \end{aligned}
\end{flalign}

Following the fused features $F$, another atrous convolution layer is implemented on top of these features to generate the segmentation mask for both the background and optimal soil sampling location categories. This process is expressed as follows:
\begin{equation}
\label{eq.mark}
M = \mbox{Conv}_{(C_m, 1)}(F),
\end{equation}
where \emph{M} is the prediction mask with dimensions of ($128 \times 128 \times 1$).

\subsubsection{Thresholding}
\begin{figure}[H]
\begin{center}
\includegraphics[width= 1\textwidth]{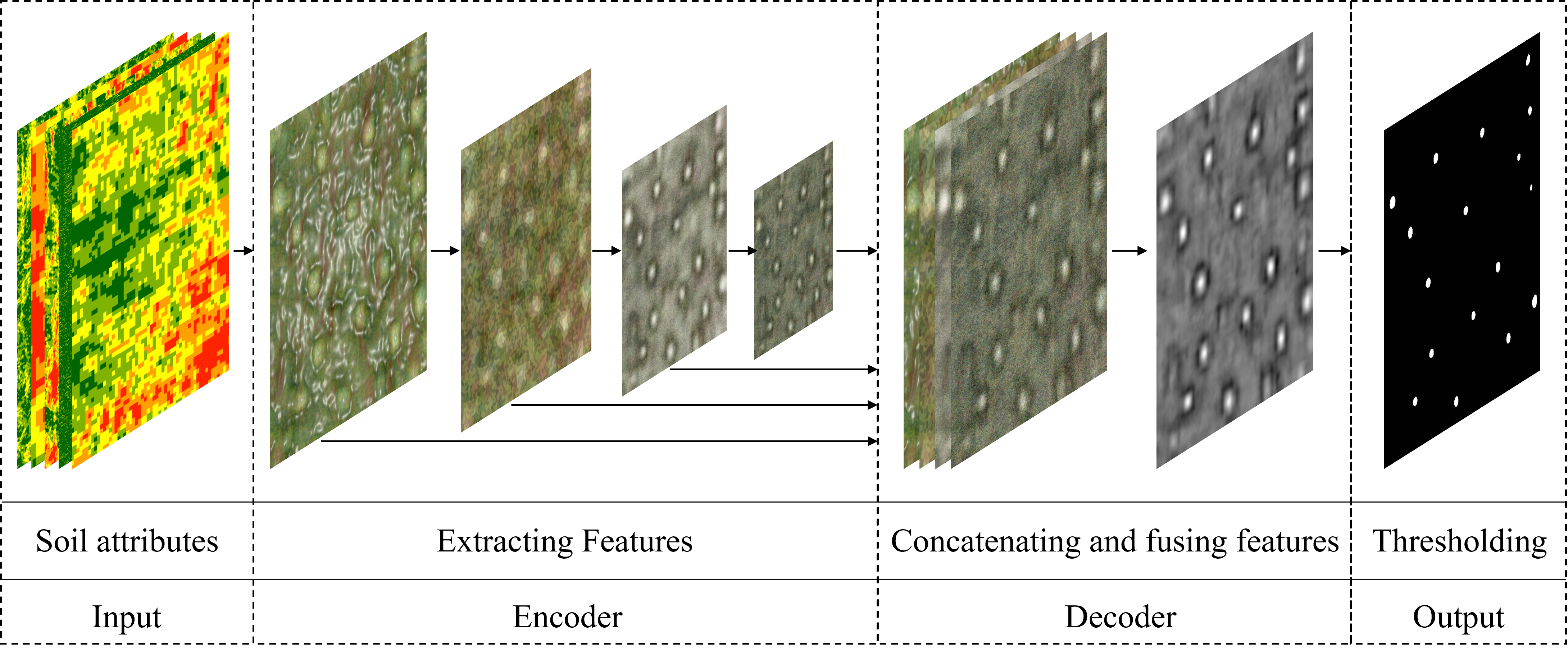}
\end{center}
\caption{General pipeline of the soil sampling site selection tool using deep learning and thresholding techniques.}
\label{fig.soil sampling site} 
\end{figure}

As previously mentioned, the prediction is a binary segmentation or a grayscale image. To obtain the final prediction, a thresholding algorithm is applied with a predefined threshold value that converts every pixel in the mask to either black pixels or white pixels, in which black pixels represent the background while white pixels indicate soil sampling locations. Consequentially, the outputs of the key components in the framework are shown in Fig. \ref{fig.soil sampling site}, and the detailed algorithm is summarized in the Algorithm \ref{algoritm}.

\begin{algorithm}[H]
	\caption{Pseudo-code explaining soil sampling site selection tool}
        \label{algoritm}
	\begin{algorithmic}[1]
        \State Input: Input images
	\For {Every set of input images (\emph{H}, \emph{W}, \emph{C})}
        \State $Tranformer$ = \lbrack \rbrack
		\For {i = 1, 2, \ldots, 4}
        		\State $F_i \leftarrow $ Shrinking input images following Equation (\ref{eq.shrinking})
            \State $F_i \leftarrow $ Embedding into sequences
            \State $F_i \leftarrow $ Modeling relationship between the input sequences by self-attention mechanism following Equations (\ref{eq. self attention}) and (\ref{eq. multi attention})
            \State $F_i \leftarrow $ Reshaping the output features to input dimensions and mixing information by $3 \times 3$ convolution
            \State Appending $F_i$ into $Tranformer$
        
		\EndFor
        \For {i = 1, 2, \ldots, 4}
        \State $\hat{F_i} \leftarrow$ Up-sampling $Tranformer_i$ to $(\frac{H}{4}\times \frac{W}{4} \times 768)$ following Equations (\ref{eq.upsample})
        \State $F \leftarrow$ Concatenating all component features $\hat{F_i}$

        \EndFor
                \State $M \leftarrow$ Fusing concatenated features $F$ $(\frac{H}{4}\times \frac{W}{4} \times 256)$ following Equation (\ref{eq.linear2}) 
                \State $M \leftarrow$ Generating prediction mask $(\frac{H}{4}\times \frac{W}{4} \times 1)$ following Equation (\ref{eq.mark})
        \State $M \leftarrow$ Up-sampling prediction $M$ to $(H \times W \times 1)$
        \State $M \leftarrow$ Applying thresholding technique for the prediction $M$

        \EndFor
        \State Computing mean Accuracy, mean Dice Coefficient, mean Intersection over Union for the predictions following Equations (14), (15), and (16)
	\end{algorithmic} 
 
\end{algorithm} 

\subsection{Training}
In the training process, parameter update is a crucial part that is conducted based on the process of optimization. The goal is to find optimal values for the model's parameters that minimize the loss between prediction and ground truth. In this work, the final outputs and ground truths are the binary images, therefore we use the binary cross-entropy loss as the optimization objective function. In particular, every pixel $i$ in the prediction has its own value $M_i$ (0 to 1). The loss function is then used to compute the difference between these pixels and their corresponding pixels on the ground truth with the corresponding value $Y_i$. Therefore, the average loss function on a pair of prediction and ground truth is designed as follows:
\begin{equation}
\label{eq.lossfunction}
\text{Average BCE Loss} (\theta) = -\frac{1}{Q} \sum_{i=1}^{Q} \Big[ Y_i \log\big(M_i(\theta)\big) + (1 - Y_i) \log\big(1 - M_i(\theta)\big) \Big],
\end{equation}
where $Q$ is the total number of pixels on the prediction or ground truth, and $i$ is the pixel's index. For example, the number of pixels $Q$ is 262144 if the dimension of the prediction is $(512 \times 512 \times 1)$. Additionally, $\theta$ contains all the parameters in our deep-learning model, and the training process adjusts these parameters to minimize the loss function.

We used stochastic gradient descent (SGD) as the primary algorithm for updating parameters as detailed in Algorithm \ref{algoritm.training}. The training process involves multiple epochs to allow the model to see the data multiple times and learn better representations. An epoch is a single pass of all the training samples through the deep learning model. In an epoch, the total number of samples $T$ is divided into multiple batches, thus to feed all samples into the model we need to iterate multiple times. The number of iterations $I$ in one epoch can be determined as
\begin{equation}
\label{eq.iteration}
I = \frac{T}{B},
\end{equation}
where $B$ is the batch size.

For each iteration, the training process is the repetition of feed-forward, loss computation, back-propagation, and parameter update. During the forward pass, the input data is fed into the model, and the model computes the predicted output. The predicted output is then compared to the ground truth labels using the loss function defined in Equation \ref{eq.lossfunction}. The loss function measures the difference between the predicted values and the ground-truth values. In the backward pass, the gradients of the loss function with respect to each parameter in the model are computed. This is done using the chain rule of calculus to propagate the errors back through the network. After obtaining the gradients, the optimization algorithm SGD updates the model's parameters to minimize the loss function. In addition, the algorithm multiplies the gradients by a learning rate $\gamma$ to control the step size for the parameter update. To prevent overfitting, we use weight decay regularization $\lambda$ in combination with the momentum $\mu$. The optimization process is summarized in Algorithm \ref{algoritm.training}.

\begin{algorithm}[H]
	\caption{Pseudo-code explaining the optimization process using SGD with momentum}
        \label{algoritm.training}
	\begin{algorithmic}[1]
        \State Input: $\gamma$ (learning rate), $\theta$ (params), $f_n(\theta)$ (objective),  $\mu$ (momentum), $\lambda$ (weight decay)

        \For {$n$ = 1 \textbf{to} \ldots $I$}
        \State $g_n \gets  \nabla_\theta f_n(\theta_{n-1})$, computing the gradient of the loss function with respect to the parameter $\theta$
        \State $g_n \gets  g_n + \lambda\theta_{n-1} $, applying regularization
        \If {$n>1$}
            \State $b_n \gets \mu b_{n-1} + g_n$, adding momentum during the training 
        \Else 
            \State $b_n \gets g_n$
        \EndIf
        \State $\theta_n \gets \theta_{n-1} - \gamma b_n$, updating the parameters

        \EndFor
        \State \textbf{return} $\theta$
	\end{algorithmic} 
        
\end{algorithm} 
\section{Performance metrics}
\label{sec.metric}
To assess the performance of the trained model, we compare the output generated by the model with the ground truth data. This allows us to compute various variables listed below that capture the differences between the model's output and the ground truth.
\begin{itemize}
\item True positive $(TP)$ is the total number of white pixels predicted from the model overlapping with the white pixels in the ground-truth data. They have the same positive category.
\item True negative $(TN)$ is the total number of black pixels predicted from the model overlapping with the black pixels in the ground-truth data. They have the same negative category.
\item False positive $(FP)$ is the total number of white pixels predicted from the model overlapping with the black pixels in the ground-truth data. They have different categories: positive prediction and negative reality.
\item False negative $(FN)$ is the total number of black pixels predicted from the model overlapping with the white pixels in the ground-truth data. They have different categories: negative prediction and positive reality.
\end{itemize}

The computed variables mentioned above, namely $(TP_i, TN_i, FP_i, \text{and} \ FN_i)$, can be utilized by comparing the model's output for each $i^{th}$ input and the ground truth dataset. To extend this comparison to the entire dataset, the segmentation performance metrics implemented in this paper are discussed in the rest of the section.

\textbf{Mean Accuracy (MA)} represents the average pixel accuracy between the ground-truth data and the model's output across all samples $T$ and is computed as follows:

\begin{equation}
\label{eq.ma}
\mbox{MA} = \frac{1}{T}\sum_{i=1}^{T}\frac{TP_i+TN_i}{TP_i+TN_i+FP_i+FN_i} 
\end{equation}
MA alone does not accurately represent the performance of the model when it comes to segmenting the minority class. Relying solely on MA can create a misleading perception that the model is performing well, even if it fails to capture the details of the minority class effectively. Therefore, we consider alternative metrics that provide a more comprehensive evaluation of the model's performance, specifically focusing on the minority class. Beside the MA metric, two other commonly used criteria to assess the effectiveness of a segmentation model are Dice Coefficient (DC) and Intersection over Union (IoU), as conducted by \cite{long2015fully} and \cite{xie2021segformer}.

\textbf{Mean Dice Coefficient (MDC)} computes the similarity between the model-generated output image and the ground truth. The quantities range from 0 to 1, and a score of 1 indicates a perfect alignment between the predicted and ground truth regions while a score of 0 signifies no overlap whatsoever. This metric is also effective in evaluating model performance on imbalanced dataset \citep{kervadec2019boundary, milletari2016v}, which applies in our case. It is defined as:
\begin{equation}
\label{eq.mdc}
\mbox{MDC} = \frac{1}{T}\sum_{i=1}^{T}\frac{2TP_i}{2TP_i+FP_i+FN_i}
\end{equation}

\textbf{Mean Intersection over Union (MIoU)} is the mean overlap of the ground truths and predictions over the total $T$ samples: 
\begin{equation}
\label{eq.iou}
\mbox{MIoU} = \frac{1}{T}\sum_{i=1}^{T}\frac{TP_i}{TP_i+FP_i+FN_i}.
\end{equation}
The IoU measures the number of white pixels that overlap between the predictions and the ground truth data.

\section{Results and discussion} 
\label{sec.resultanddiscussion}
In this section, we compare and discuss the results of the model we designed with respect to the state-of-the-art CNN-based segmentation model proposed by \cite{acharya2023deep}. 

\subsection{Parameter setting}
\label{param}
For the training process, all models are implemented based on the deep learning framework PyTorch 1.13.1 and the HuggingFace library. In our model, we replaced the SegFormer decoder with our designed atrous convolution networks to improve the prediction accuracy and robustness as described in the methodology section. 

In order to train the model, we use a high-performance computing facility, called AI.Panther, equipped with A100 SXM4 GPUs at the Florida Institute of Technology. Before calculating the metrics described in Section \ref{sec.metric}, we perform a post-processing step on the output images obtained from the trained model. This involves scaling the pixel values to a range of 0 to 255 and then applying a thresholding technique using a standard threshold value of 190. Any pixel value below 190 is considered 0, representing the background, while pixel values equal to or above 190 are set to 255, indicating the optimal soil sampling locations.

The model is trained with a batch size $B$ of 4 and the BCE loss. To aid in optimization, we used the learning rate $\gamma$ of 0.001 incorporating a momentum value $\mu$ of 0.9. To mitigate over-fitting, a weight decay $\lambda$ of 0.0001 was applied to the optimization algorithm. The decision to use a small batch size was driven by hardware limitations. When a larger batch size was attempted, Nvidia CUDA memory errors occurred. The performance results accomplished by our proposed method with be discussed in detail in the next section.

\subsection{This paper's method}
\begin{figure}[H]
\begin{center}
\includegraphics[width= 0.495\textwidth]{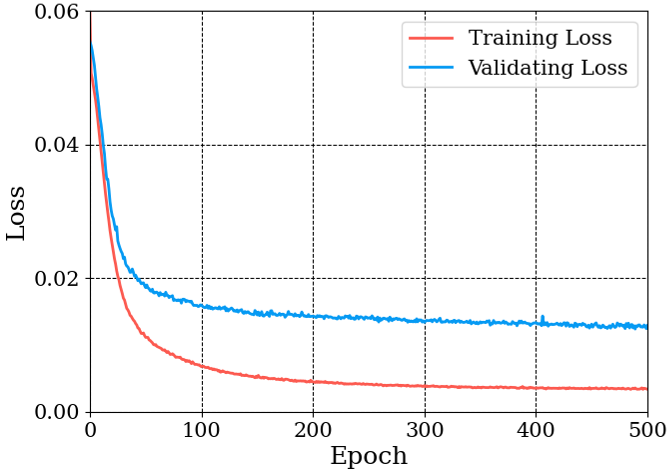}
\end{center}
\caption{The total loss with respect to the total number of epochs in the training process of our model.}
\label{fig.Loss_atrous} 
\end{figure}

The training progress of our method is presented in Fig. \ref{fig.Loss_atrous}. As seen in Fig. \ref{fig.Loss_atrous}, the training loss is converged after 500 epochs. However, it is important to note that a training loss value of less than 0.01 does not directly guarantee a model accuracy of 99\%. As discussed, the loss value can be low even if the model outputs all pixels as black due to unbalanced data (the number of black pixels is substantially larger than the number of white pixels). Therefore, we compute various performance metrics using the trained model to assess its true effectiveness.

To gain a deeper understanding of the overall training behavior of the model, including overfitting and undertraining, performance metrics discussed in Section~\ref{sec.metric} are computed for our model and saved every 50 epochs during the training process. The saved model is loaded, and performance metrics are computed on the training, validation, and testing datasets. If the performance metrics on the validation and testing datasets decrease over time, it suggests that the model may have been overtrained. Additionally, if the performance metrics of the model continue to increase as the number of epochs increases, it might be beneficial to continue training for additional epochs. Therefore, the performance of the trained model, along with the model's training and validation loss, is utilized to evaluate the model's learning ability more comprehensively. 

\begin{figure}[h]
  \centering
  \begin{subfigure}{0.49\textwidth}
    \centering
    \includegraphics[width=\linewidth]{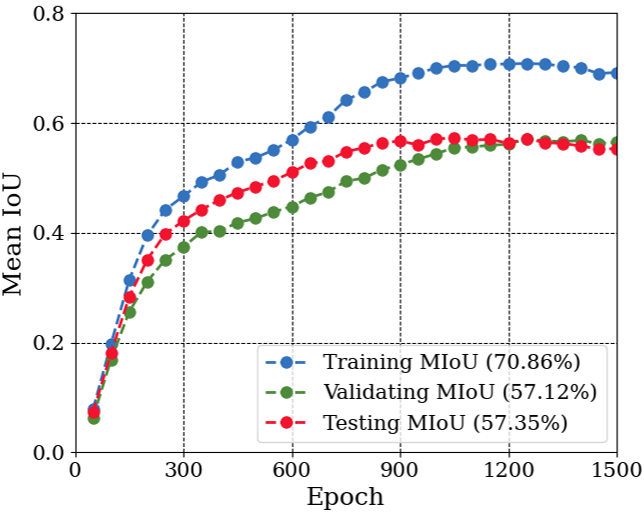}
    \caption{Mean Intersection over Union.}
    \label{fig.iou_atrous}
  \end{subfigure}
\hfill
  \begin{subfigure}{0.49\textwidth}
    \centering
    \includegraphics[width=\linewidth]{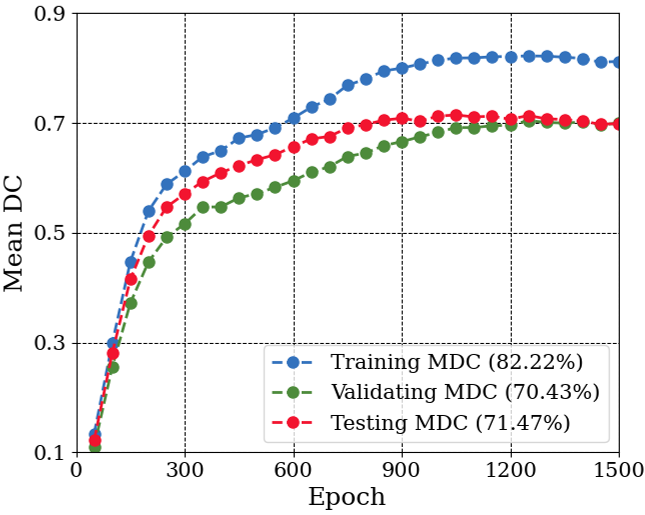}
    \caption{Mean Dice Coefficient.}
    \label{fig.dice_atrous}
  \end{subfigure}
  \vspace{1em}
  \begin{subfigure}{0.49\textwidth}
    \centering
    \includegraphics[width=\linewidth]{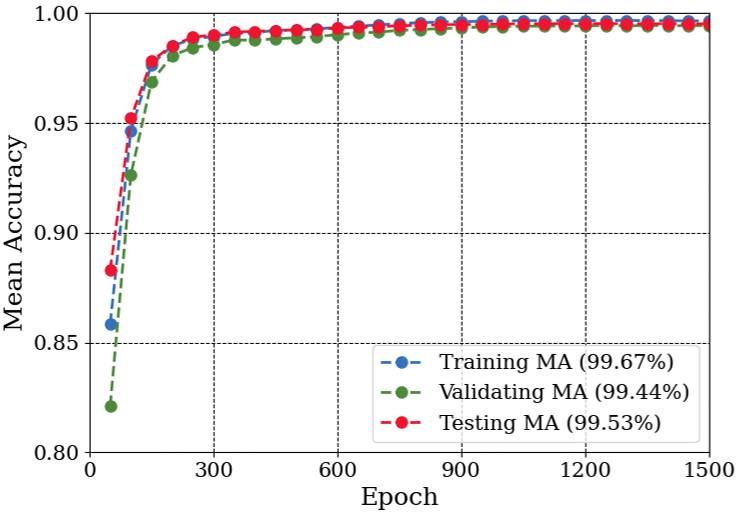}
    \caption{Mean accuracy.}
    \label{fig.acc_atrous}
  \end{subfigure}
  \caption{Performance metrics computed by using our model on training, validation, and testing data.}
  \label{fig:results_atrousSegformer}
\end{figure}

Figures \ref{fig.iou_atrous}, \ref{fig.dice_atrous}, and \ref{fig.acc_atrous} show the performance metrics values computed using the trained model at different iterations. In these figures, we only show the training progress up to 1500 epochs because the MIoU and MDC deteriorate after 1200 epochs on the train, validation, and test datasets. The highest MA,  MIoU, and MDC on the test dataset are 99.53\%, 57.35\%, and 71.47\%, respectively. Additionally, the highest performance values are achieved when the model training reaches approximately 1050 epochs on the test dataset. A similar conclusion can be drawn when examining the training and validation loss in Fig. \ref{fig.Loss_atrous} as both loss starts to saturate at around 1050 epoch. Thus, we pick the model saved at 1050 epochs as the best model during our overall training phase and use this model to carry out inference on the test dataset. 

\begin{figure}[h]
    \centering 
\begin{subfigure}{0.49\textwidth}
  \includegraphics[width=\linewidth]{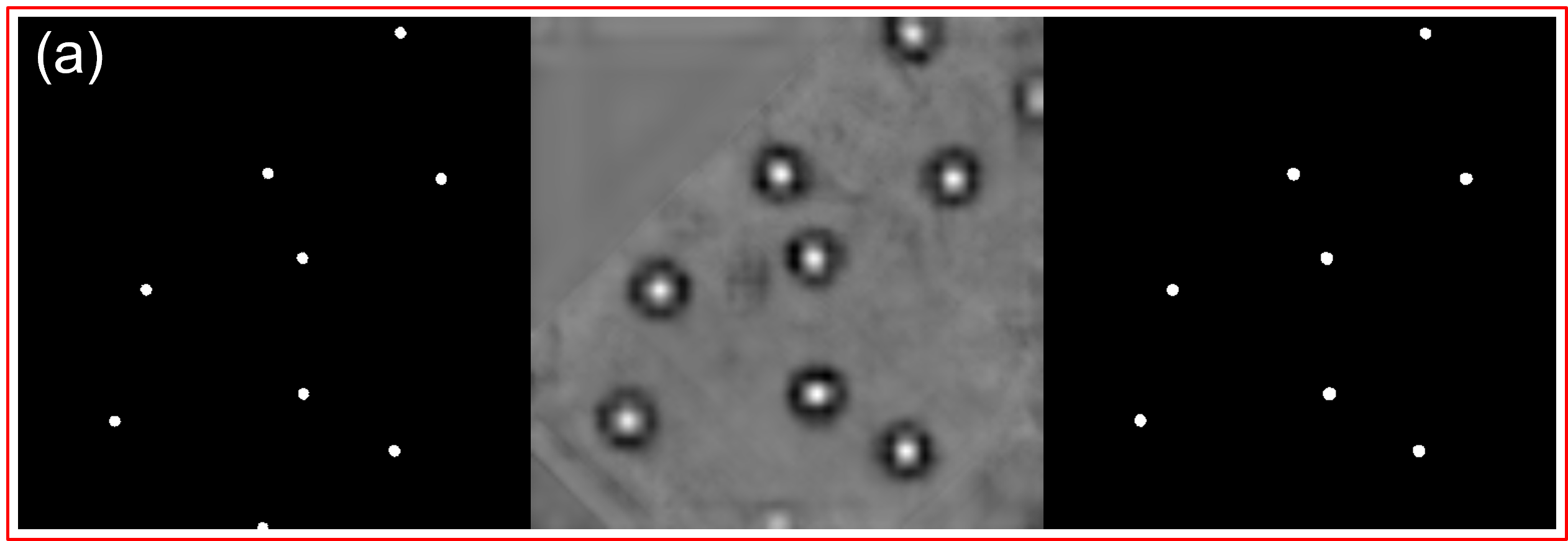}
\end{subfigure}\hfil 
\begin{subfigure}{0.49\textwidth}
  \includegraphics[width=\linewidth]{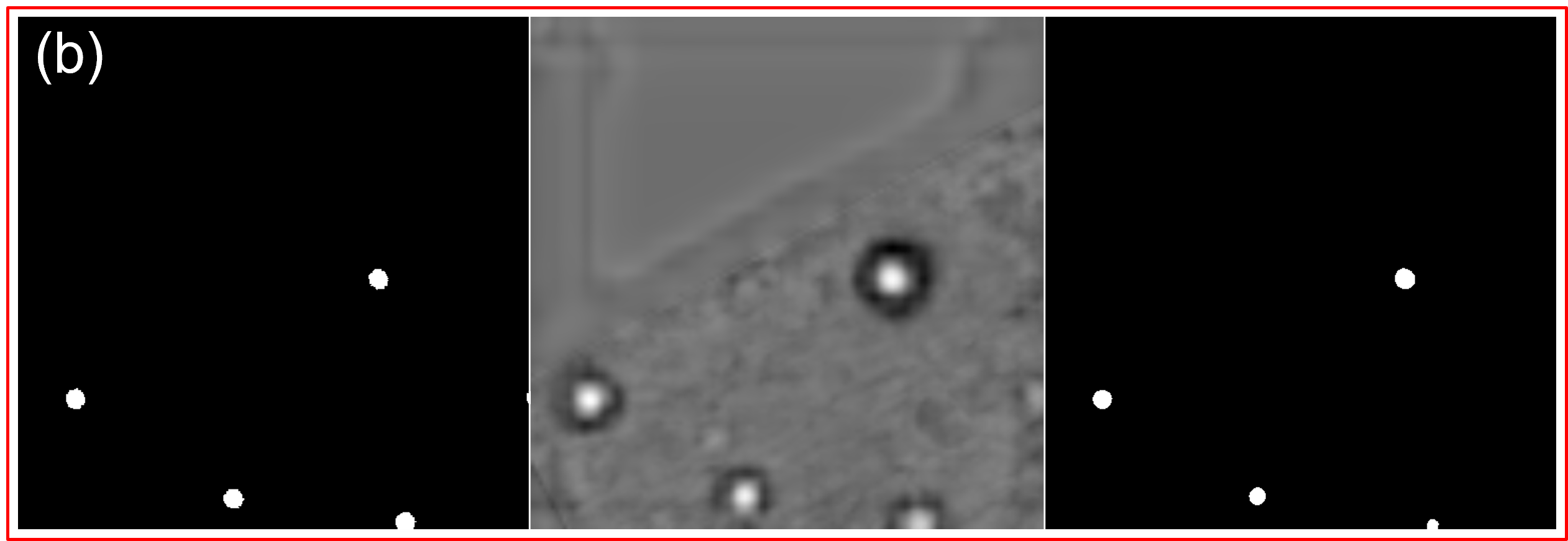}
\end{subfigure}\hfil 
\medskip

\begin{subfigure}{0.49\textwidth}
  \includegraphics[width=\linewidth]{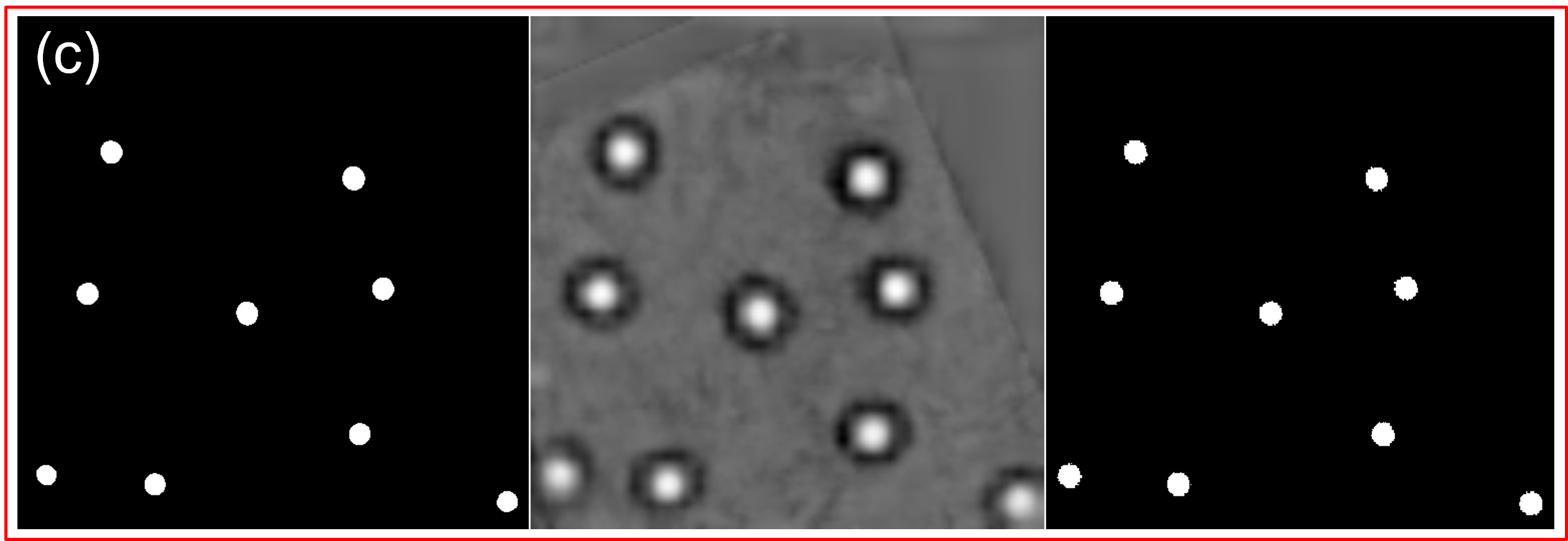}
\end{subfigure}\hfil 
\begin{subfigure}{0.49\textwidth}
  \includegraphics[width=\linewidth]{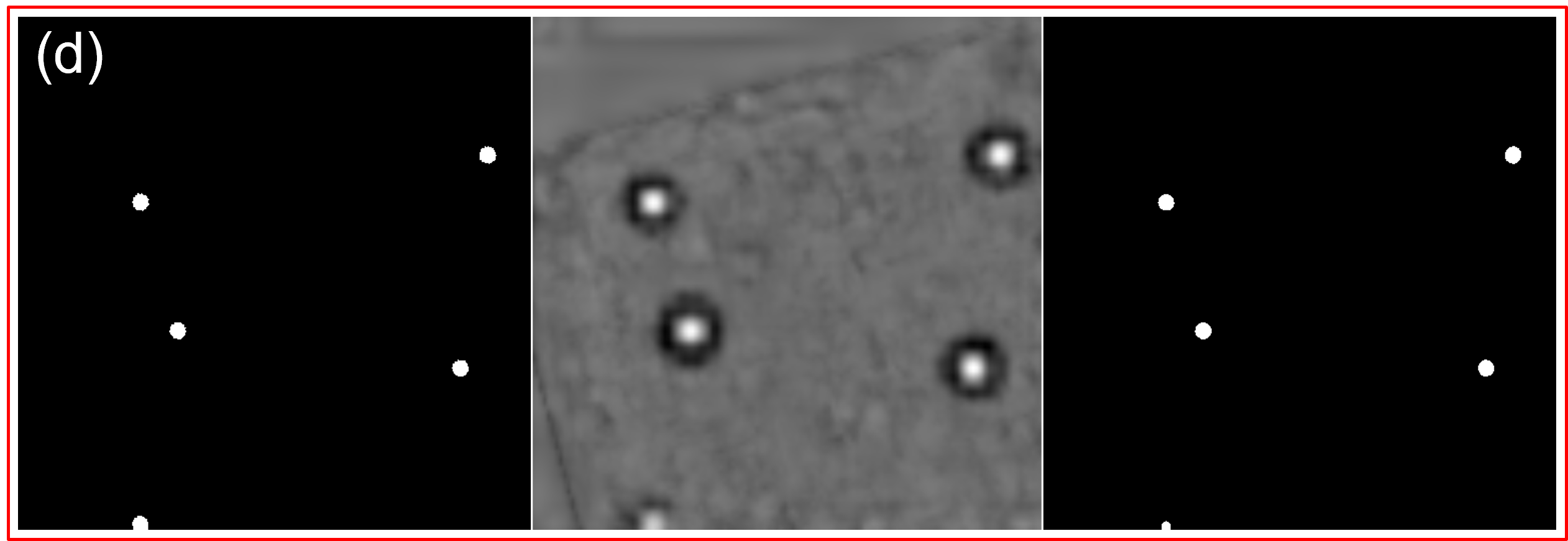}
\end{subfigure}\hfil 
\medskip

\begin{subfigure}{0.49\textwidth}
  \includegraphics[width=\linewidth]{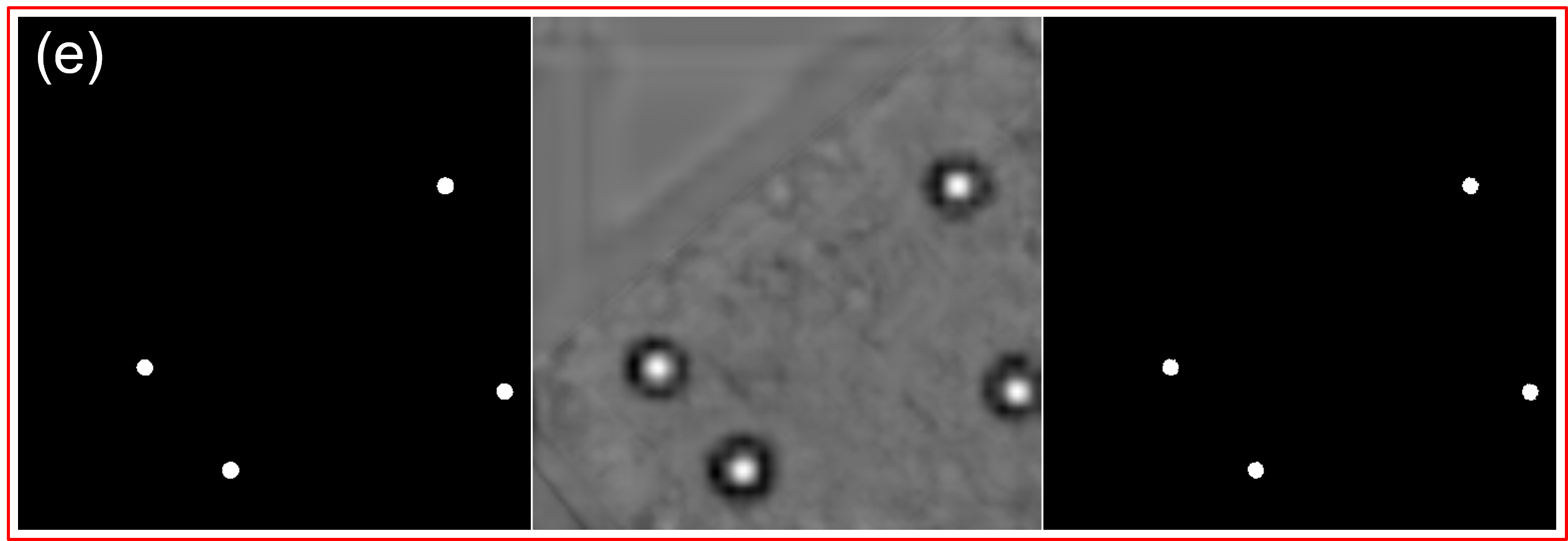}
\end{subfigure}\hfil 
\begin{subfigure}{0.49\textwidth}
  \includegraphics[width=\linewidth]{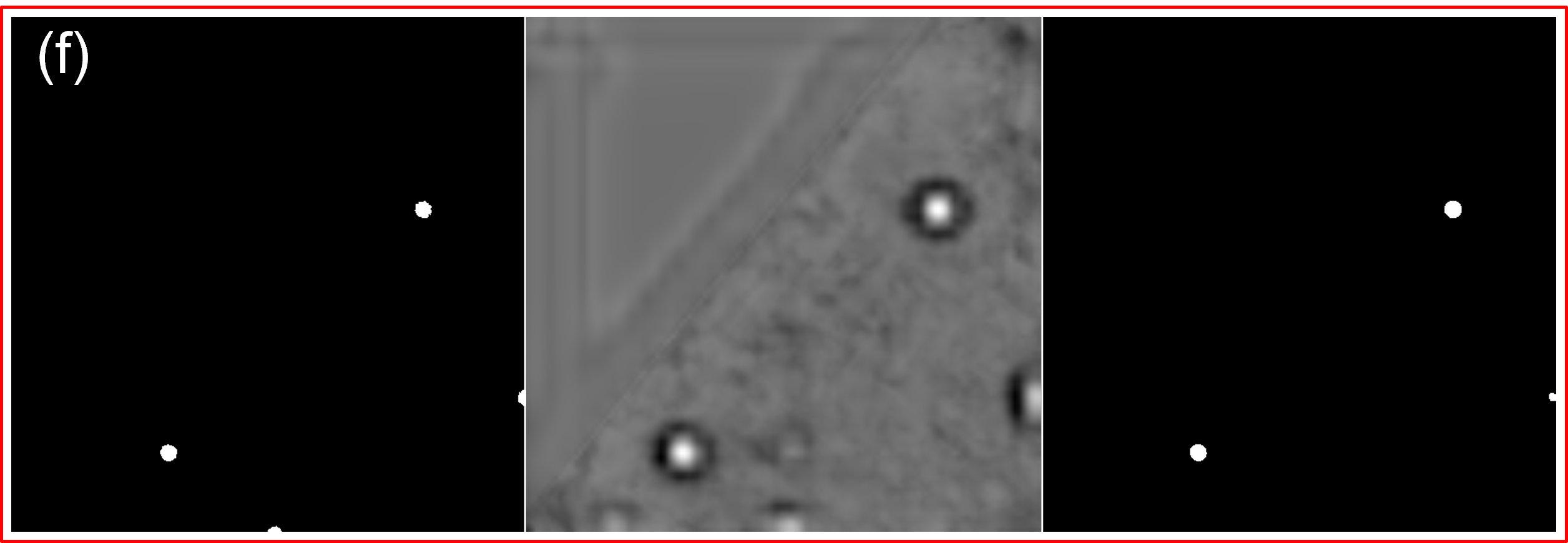}
\end{subfigure}\hfil 
\medskip

\begin{subfigure}{0.49\textwidth}
  \includegraphics[width=\linewidth]{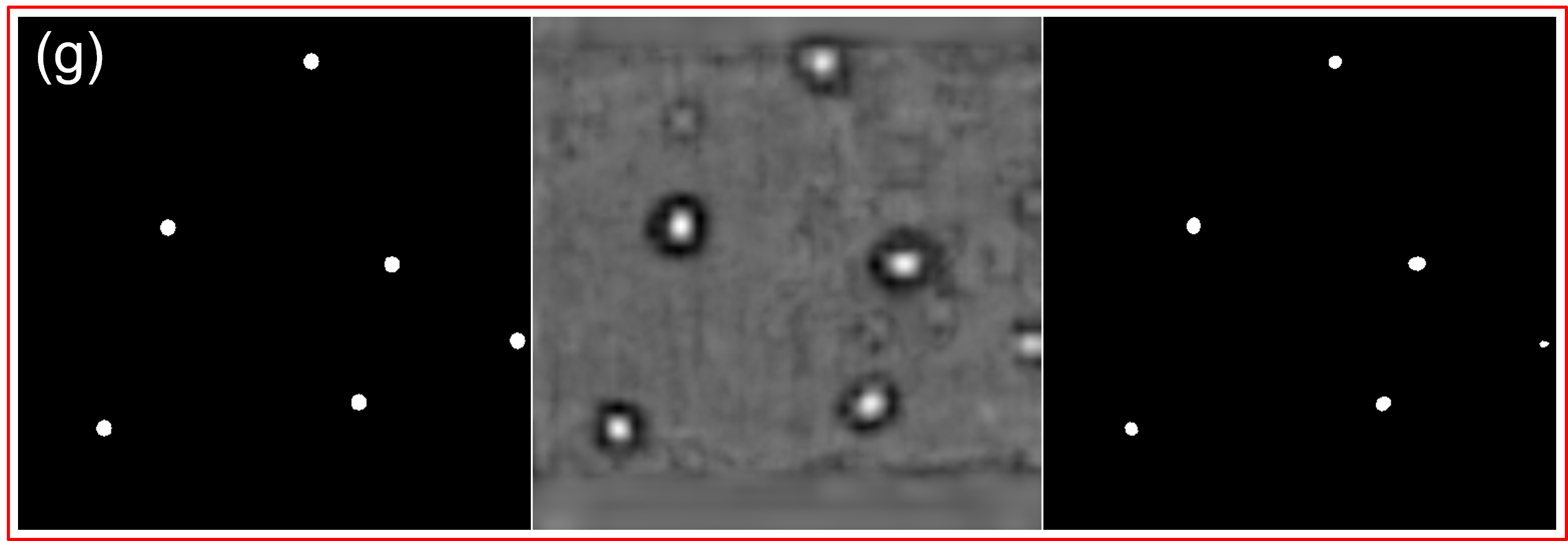}

\end{subfigure}\hfil 
\begin{subfigure}{0.49\textwidth}
  \includegraphics[width=\linewidth]{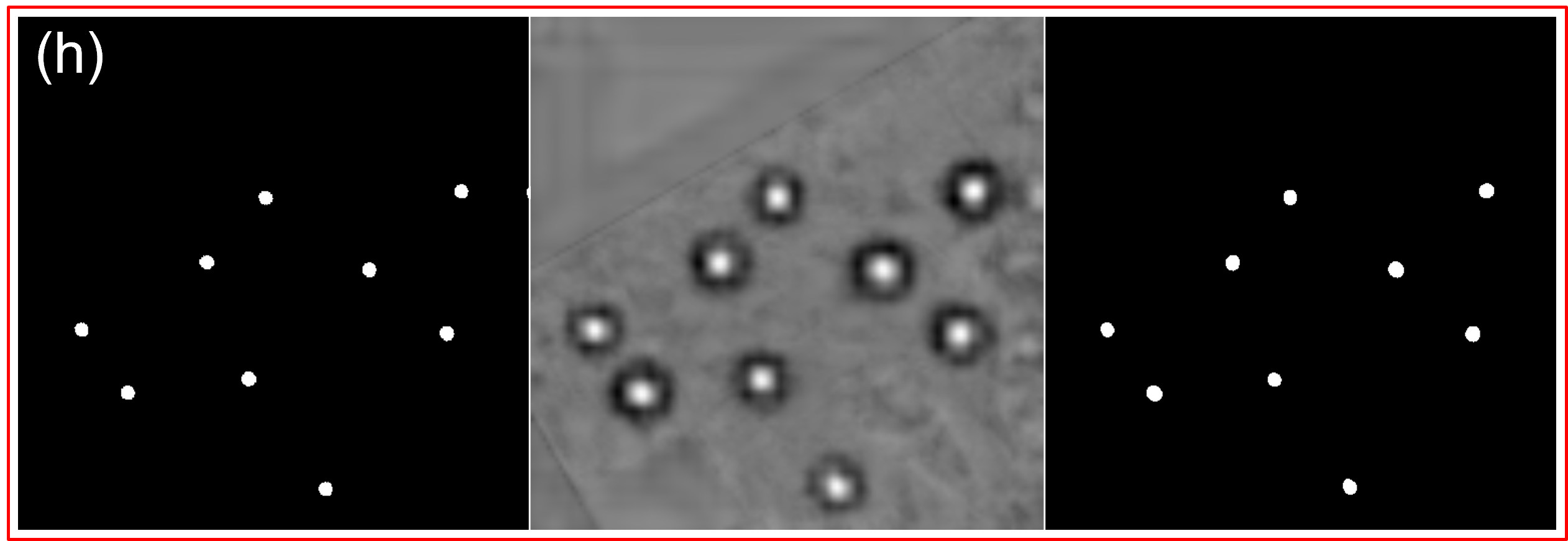}

\end{subfigure}\hfil 
\caption{Randomly representative results predicted from the model on the testing dataset. In a typical sub-figure (a-h), the ground truth, the predicted result corresponding to the ground truth, and the threshold of the predicted result are shown on the left, center, and right, respectively.}
\label{fig.result}
\end{figure}

Figure~\ref{fig.result} shows the performance of the model on the test dataset for eight different fields. Each result panel (a-h) represents a different field. In each panel, the ground truth, the predicted soil sampling sites, and the predicted result after thresholding are shown on the left, center, and right, respectively. Comparing the model's predictions with the ground truths, we can see that the model successfully predicts the majority of the essential soil sampling sites within the given fields. However, it is worth noting that the predicted data contain background areas that are not purely black. Therefore, a thresholding technique is applied to ensure clarity and facilitate metric computation.

\subsection{CNN-based Segmentation Method}
\label{pre_work}
This section discusses the results when a CNN-based model is designed and implemented to solve the soil sampling problem and compares its performance with our proposed model.  After the training data is prepared same as the previous sections, The CNN-based framework proposed in research work \citep{acharya2023deep} was trained and tested for the task of soil sampling site segmentation. However, during the initial analysis, the success observed for crop-spray systems in \citep{acharya2023deep} could not be replicated for the soil sampling problem. This discrepancy prompted further investigation into the possible causes of the problem. One of the primary challenges encountered by the deep-learning model is directly associated with the characteristics of the data itself: pixel class imbalance.

One such solution involved the utilization of the focal loss function, which has been proposed and used in numerous applications, including, but not limited to, studies by \citep{panella2022semantic,jadon2020survey,yeung2022unified,chang2018brain}, where pixel class imbalance was a primary concern. Another aspect explored during the investigation was converting input images to grayscale for model training. The hypothesis assumed that utilizing the grayscale representation of each attribute would be sufficient for the model to segment the potential soil sampling locations. Building upon these understandings, an experimental setup was designed with the following variations:

\begin{itemize}
    \item The input images were converted to grayscale images, and binary cross entropy (BCE) loss was used.
    \item The input images were converted to grayscale, and focal loss was used.
    \item RGB input images were used with BCE loss.
    \item RGB input images were used with Focal loss.
\end{itemize}

We conducted an experiment using the CNN-based model, throughout this experiment, we employed specific hyper-parameters. The batch size was set to 32 while the other hyperparameters such as learning rate and optimizer remain the same as presented in Section~\ref{param}. Furthermore, the progress of the trained model was saved every 500 epochs, ensuring consistent checkpoints for evaluation and comparison.

\begin{figure}[H]
    \centering    \includegraphics[width=\linewidth]{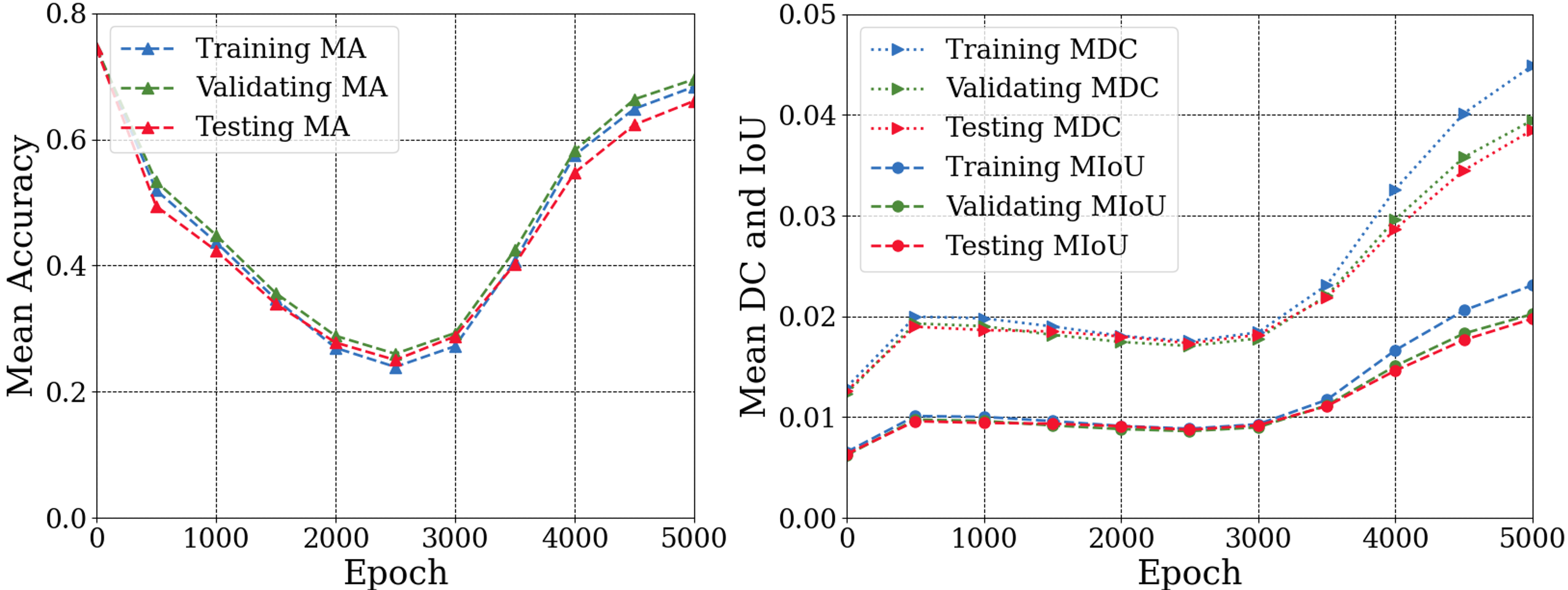}
    \caption{ Grayscale images with binary cross entropy loss.}
    \label{fig:gray_bce}
\end{figure}

\begin{figure}[H]
    \centering    \includegraphics[width=\linewidth]{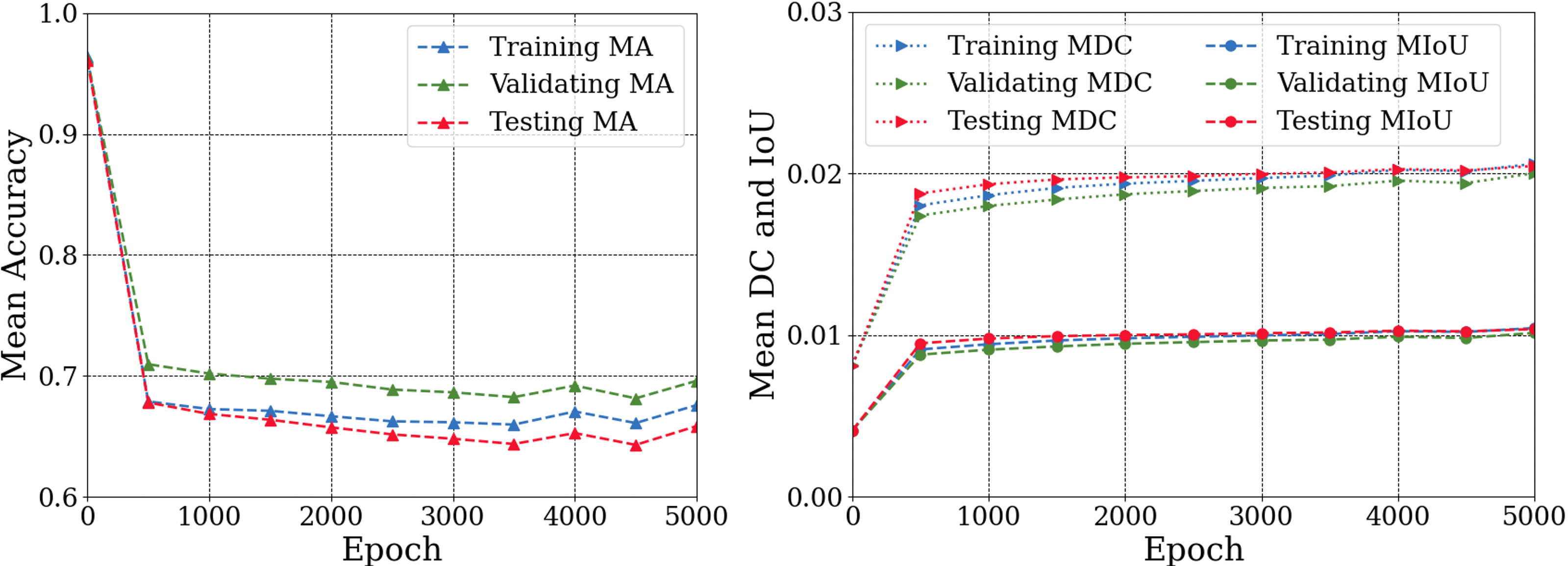}
    \caption{Grayscale images with focal loss.}
    \label{fig:gray_focal}
\end{figure}

\begin{figure}[H]
    \centering    \includegraphics[width=\linewidth]{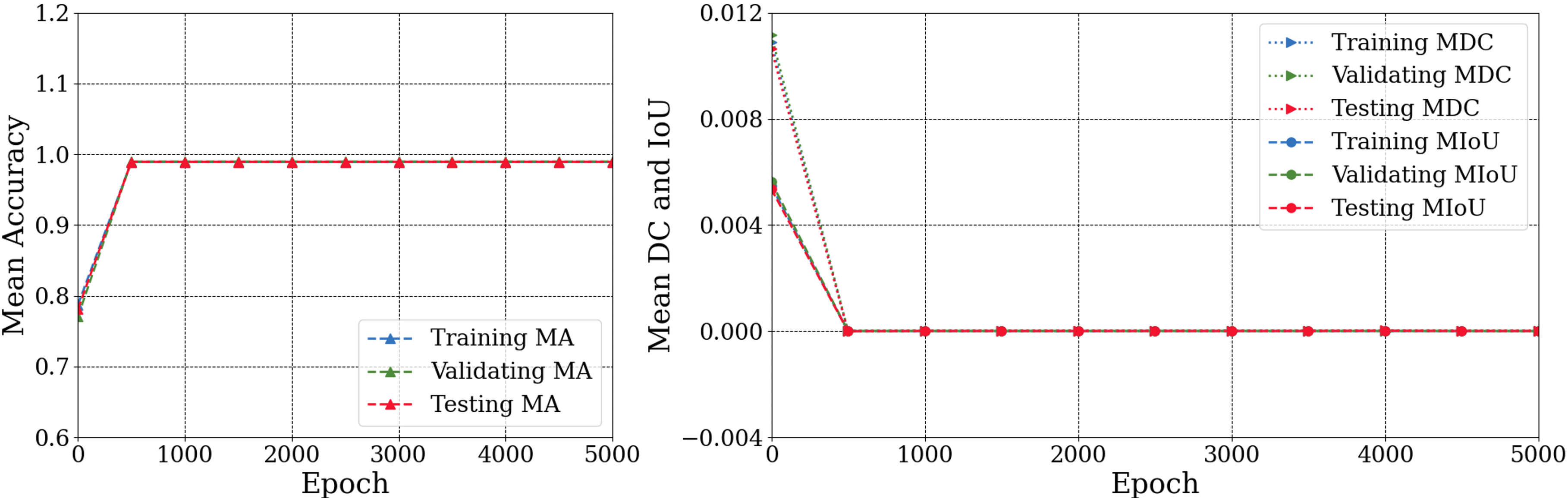}
    \caption{RGB input images with binary cross entropy loss.}
    \label{fig:rgb_bce}
\end{figure}

\begin{figure}[H]
    \centering    \includegraphics[width=\linewidth]{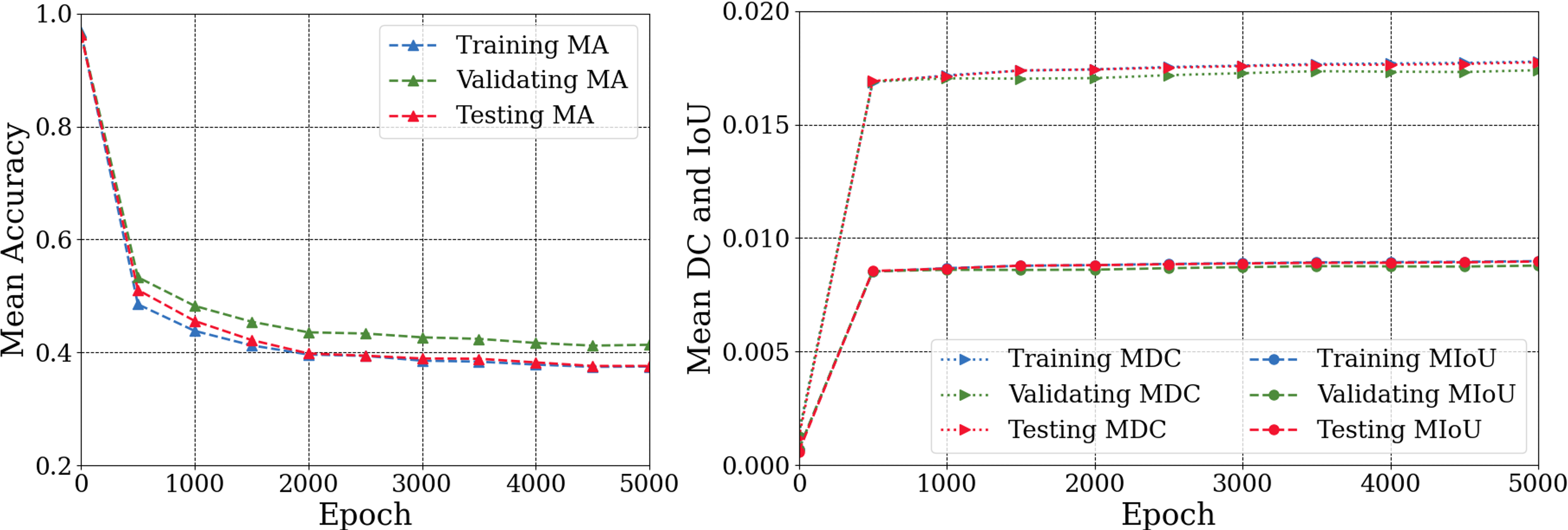}
    \caption{RGB input images with focal loss.}
    \label{fig:rgb_focal}
\end{figure}

Figures \ref{fig:gray_bce} - \ref{fig:rgb_focal} offer insights into the CNN-based model's performance evaluation using the same metrics. Despite training the model for 5000 epochs, it fails to converge. Figures \ref{fig:gray_bce} and \ref{fig:gray_focal} depict the results when using grayscale input images with the BCE loss and Focal loss, respectively. On the other hand, Figures \ref{fig:rgb_bce} and \ref{fig:rgb_focal} showcase the outcomes when utilizing RGB input images with BCE loss and Focal loss, respectively. 

The results of the CNN-based model indicate that, at the 5000th epoch, the MA of the model trained with BCE loss consistently outperforms the model trained with Focal loss. However, this superiority in accuracy does not necessarily guarantee better performance across other metrics. There is no significant difference observed between the models trained with Focal loss and BCE loss in terms of MDC and MIoU. This experiment, which involved various variations in terms of the loss function and model inputs, unfortunately, resulted in poor performance. The consistent observation of these results led us to delve deeper into our investigation by exploring more complex models, particularly those based on the transformer architecture as discussed in Section \ref{sec.3}.

\subsection{Comparison Summary}
In summary, Table \ref{table.comparison} lists the key performance metrics for the two methods in comparison: the model we designed in this paper and the CNN-based model. We can see that our model outperforms the other on all train, validation, and test datasets.

\begin{table}[H]
\centering
\caption{Performance comparison of our model with respect to the CNN-based model on the train, validation, and test datasets.}
\label{table.comparison}
\begin{tabular}{l@{\hspace{6pt}}ccc@{\hspace{5pt}}|@{\hspace{5pt}}ccc@{\hspace{5pt}}|@{\hspace{5pt}}cccc}
\toprule
\multirow{2}{*}{Methods} & \multicolumn{3}{c}{Train} & \multicolumn{3}{c}{Validation} & \multicolumn{3}{c}{Test}\\
\cline{2-11}  \\[-1em]
& MA & MDC & MIoU & MA & MDC & MIoU & MA & MDC & MIoU& \\

\midrule  \addlinespace[-0.25ex] \midrule
Our model & 99.66 & 81.83 & 70.5 & 99.45 & 70.43 & 57.12 & 99.52 & 71.47 & 57.35 & \\ 
\midrule
\begin{tabular}{@{}c@{}}CNN-based \\ model \end{tabular} & 68.34 & 4.49 & 2.31 & 69.51 & 3.95 & 2.02 & 66.08 & 3.85 & 1.98& \\  
\bottomrule
\end{tabular}
\end{table}

\section{Conclusion}
\label{sec.6}
This work developed a tool for soil-sampling location selection to optimize the soil-sampling process. The main challenge is the imbalanced dataset, which significantly affects the task of developing the model to identify soil sampling sites. To tackle this issue, we initially approached a CNN-based model built on top of the Unet architecture. We conducted experiments on the CNN-based model with binary or RGB input images in combination with Binary Cross Entropy or Focal loss as the loss functions. However, the CNN-based model did not show its potential for building a soil sampling tool as reported in our findings. This leads us to develop models with transformers equipped with self-attention as the backbone of the tool to better handle the imbalanced dataset and yield improved results. 

We designed our model with self-attention as the main mechanism to capture and model the relationships between different patches in an input dataset. This mechanism allows the model to effectively extract relevant features from the input dataset. Then, these features are upsampled, concatenated, and fused using multi-layer perceptron and atrous convolution networks. The output generated by the model passes to a post-processing operation, which involves using a thresholding technique. A threshold value of 190 is applied to the output prediction to generate maps with black pixels indicating the background and white pixels representing the optimal soil-sampling locations. The post-processing process exports the locations corresponding to the soil sampling sites within a given field.

For experimental validation, we demonstrated that our model outperforms the state-of-the-art CNN-based model on the soil sampling dataset. Our model achieves 99.52${\%}$ mean accuracy, 57.35${\%}$ mean IoU, and 71.47${\%}$ mean Dice Coefficient on the testing dataset while those values of the CNN-based model are 66.08\%, 3.85\%, and 1.98\%, respectively. To the best of our knowledge, our research represents the first work that utilizes deep learning techniques to enable automated soil-sampling location selection. The tool can significantly enhance soil sampling and analysis, thereby advancing our understanding of soil health.

In future work, we will continue to investigate the framework to further improve the performance and disseminate it to producers in the form of a mobile app with GPS guidance to go from spot to spot. The tool will also allow for the mixing of appropriate samples and provide information on the estimated cost of analysis. If the producer chooses a price cap, the tool could inform them of the number of samples that could be analyzed within that price range and how accurate the results would be.

\section{Acknowledgment}
This work is supported by grants ${\#}$2021-67022-38910 and ${\#}$2022-67021-38911 from the USDA National Institute of Food and Agriculture.
\label{sec.acknow}

\bibliographystyle{elsarticle-harv} 
\bibliography{reference}





\end{document}